\documentclass[lettersize,journal]{IEEEtran}
\usepackage{amsmath,amsfonts}
\usepackage{algorithmic}
\usepackage{algorithm}
\usepackage{array}
\usepackage[caption=false,font=normalsize,labelfont=sf,textfont=sf]{subfig}
\usepackage{textcomp}
\usepackage{stfloats}
\usepackage{url}
\usepackage{verbatim}
\usepackage{graphicx}
\usepackage{cite}
\usepackage{multirow}
\usepackage{hyperref}
\usepackage[table]{xcolor}
\usepackage{booktabs}

\usepackage{soul}

\newcommand{\old}[1]{{\color{blue}{#1}}} 
\newcommand{\new}[1]{{\color{black}{#1}}} 
\newcommand{\boldparagraph}[1]{\vspace{0.2cm}\noindent{\bf #1:} }
\definecolor{top1}{HTML}{ff9999}
\definecolor{top2}{HTML}{ffcc99}
\definecolor{top3}{HTML}{fff6b2}

\begin{document}

\title{LM-Gaussian: Boost Sparse-view 3D Gaussian Splatting with Large Model Priors}
% \title{Supplementary Material for LM-Gaussian}
\author{Hanyang Yu$^{1}$,
        Xiaoxiao Long$^{1,\ddag}$
        and Ping Tan$^{1}$
        \thanks{\IEEEcompsocthanksitem $^\ddag$ Xiaoxiao Long is the corresponding author (xxlong@connect.hku.hk).
\IEEEcompsocthanksitem $^{1}$ The Hong Kong University of Science and Technology }}
% \IEEEcompsocthanksitem Submitted for review on May. 6th, 2025.}}

\markboth{Journal of \LaTeX\ Class Files,~Vol.~14, No.~8, May~2025}
{Shell \MakeLowercase{\textit{et al.}}: A Sample Article Using IEEEtran.cls for IEEE Journals}

\maketitle

\begin{abstract}
We aim to address sparse-view reconstruction of a 3D scene by leveraging priors from large-scale vision models. While recent advancements such as 3D Gaussian Splatting (3DGS) have demonstrated remarkable successes in 3D reconstruction, these methods typically necessitate hundreds of input images that densely capture the underlying scene, making them trivial and sometimes impractical for real-world applications. However, sparse-view reconstruction is inherently ill-posed and under-constrained, often resulting in inferior and incomplete outcomes. This is due to issues such as failed initialization, overfitting on input images, and a lack of details.
To mitigate these challenges, we introduce LM-Gaussian, a method capable of generating high-quality reconstructions from a limited number of images. Specifically, we propose a robust initialization module that leverages stereo priors to aid in the recovery of camera poses and the reliable point clouds. Additionally, a diffusion-based refinement is iteratively applied to incorporate image diffusion priors into the Gaussian optimization process to preserve intricate scene details. Finally, we utilize video diffusion priors to further enhance the rendered images for realistic visual effects.
% 3D Gaussian Splatting (3DGS) has found widespread applications in the field of 3D reconstruction. However, vanilla 3DGS heavily relies on the initialization of point clouds and camera poses from structure from motion (SFM) techniques, often requiring hundreds of images to reconstruct a scene. It's hard for people to obtain such dense images in reality, raising bar for this technology and causing inconvenice. 
% Given that, we introduce $LM-Gaussian$, a novel coarse-to-fine framework to $Boost$ $Sparse$ $view$ $3D$ $Gaussian$ $Splatting$
% $with$ $Large$ $Model$ $Priors$, which cleverly integrate the strengths of different large model priors: two-view initialization model, monocular estimation model and latent diffusion model, enabling sparse-view reconstruction of an entire 360-degree scene. In this work, we demonstrate how ingenious fusion of priors from large models can aid in the process of scene reconstruction.  To unleash the power of these priors, we use multi-modal prior-guided initialization to get dense point clouds and poses. Then we adopt a coarse-to-fine training strategy. First we use depth, normal and virtual views to regularize gaussian optimization. Then we take advantage of diffusion prior, which can restore real-world details, to refine gaussian reconstruction. 
Overall, our approach significantly reduces the data acquisition requirements compared to previous 3DGS methods. We validate the effectiveness of our framework through experiments on various public datasets, demonstrating its potential for high-quality 360-degree scene reconstruction. Visual results are on our website\href{https://runningneverstop.github.io/lm-gaussian.github.io/}{(lm-gaussian.github.io)}, and the code has been released at \href{https://github.com/hanyangyu1021/LMGaussian}{(https://github.com/hanyangyu1021/LMGaussian)}
\end{abstract}

\begin{IEEEkeywords}
sparse-view, reconstruction, gaussian splatting, large models.
\end{IEEEkeywords}

\section{Introduction}

3D scene reconstruction and novel view synthesis from sparse-view images present significant challenges in the field of computer vision. Recent advancements in neural radiance fields (NeRF) \cite{mildenhall2020nerf} and 3D Gaussian splatting (3DGS) \cite{kerbl20233d} have made notable progresses in synthesizing novel views, but they typically require hundreds of images to reconstruct a scene. Capturing such a dense set of images is often impractical, raising the inconvenience for utilizing these technologies. Although efforts have been made to address sparse-view settings, existing works are still limited to straightforward facing scenarios, such as the LLFF dataset \cite{mildenhall2019local}, which involve small-angle rotations and simple orientations. For large-scale 360-degree scenes, the problems of being ill-posed and under-constrained hinder the employment of these methods. In this work, we present a new method that is capable of producing high-quality reconstruction from sparse input images, demonstrating promising results even in challenging 360-degree scenes.

There are three main obstacles that prevent 3D Gaussian splatting from achieving high-quality 3D reconstruction with sparse-view images. 1) \textbf{Failed initialization:} 3DGS heavily relies on pre-calculated camera poses and point clouds for initializing Gaussian spheres. However, traditional Structure-from-Motion (SfM) techniques \cite{schonberger2016pixelwise} cannot successfully handle the sparse-view setting due to insufficient overlap among the input images, therefore yielding inaccurate camera poses and unreliable point clouds for 3DGS initialization. 2) \textbf{Overfitting on input images:} Lacking sufficient images to provide constraints, 3DGS tends to be overfitted on the sparse input images and therefore produces novel synthesized views with severe artifacts. 3) \textbf{Lack of details:} Given limited multi-view constraints and geometric cues, 3DGS always fails to recover the details of the captured 3D scene and the unobserved regions, which significantly degrades the final reconstruction quality.

To tackle these challenges, we introduce LM-Gaussian, a novel method capable of producing high-quality reconstructions from sparse input images by incorporating large model priors. The key idea is leveraging the power of various large model priors to boost the reconstruction of 3D gaussian splatting with three primary objectives: 1) \textbf{Robust initialization}; 2) \textbf{Overfitting prevention}; 3) \textbf{Detail preservation}.

For robust initialization, instead of relying on traditional SfM methods \cite{schonberger2016structure,schonberger2016pixelwise}, we propose a novel initialization module utilizing stereo priors from DUSt3R \cite{wang2024dust3r}. DUSt3R is a stereo model that takes pairs of images as input and directly generates corresponding 3D point clouds. Through a global optimization process, it derives camera poses from the input images and establishes a globally registered point cloud.
However, the global point cloud often exhibits artifacts and floaters in background regions due to the inherent bias of DUSt3R towards foreground regions.
To mitigate this issue, we introduce a Background-Aware Depth-guided Initialization module. Initially, we use depth priors to refine the point clouds produced by DUSt3R, particularly in the background areas of the scene. Additionally, we employ iterative filtering operations to eliminate unreliable 3D points by conducting geometric consistency checks and confidence-based evaluations. This approach ensures the generation of a clean and reliable 3D point cloud for initializing 3D Gaussian splatting.

Once a robust initialization is obtained, photo-metric loss is commonly used to optimize 3D Gaussian spheres. 
However, in the sparse-view setting, solely using photo-metric loss will make 3DGS overfit on input images. To address this issue, we introduce multiple geometric constraints to regularize the optimization of 3DGS effectively.
Firstly, a multi-scale depth regularization term is incorporated to encourage 3DGS to capture both local and global geometric structures of depth priors. Secondly, a cosine-constrained normal regularization term is introduced to ensure that the geometric variations of 3DGS to be aligned with normal priors. Lastly, a weighted virtual-view regularization term is applied to enhance the resilience of 3DGS to unseen view directions.

To preserve intricate scene details, we introduce Iterative Gaussian Refinement Module, which leverages diffusion priors to recover high-frequency details. 
We leverage a diffusion-based Gaussian repair model to restore the images rendered from 3DGS, aiming to enhance image details with good visual effects. The enhanced images are used as additional pseudo ground-truth to optimize 3DGS. Such a refinement operation is iteratively employed in 3DGS optimization, which gradually inject the image diffusion priors into 3DGS for detail enhancement.
Specifically, the Gaussian repair model is built on ControlNet with injected Lora layers, where the sparse input images are used to finetune the Lora layers so that repair model could work well on specific scenes.

% Initially, we train a Gaussian Repair Model by fine-tuning a ControlNet with injected Lora weights. This model is adept at repairing Gaussian-rendered images, resulting in outputs rich in detail. Subsequently, an iterative Gaussian optimization approach is utilized for scene refinement. To ensure viewpoint consistency and scene authenticity, we systematically reintroduce details to the Gaussian-rendered images, using the repaired images as guides throughout the Gaussian optimization process. Following a specified number of iterations denoted by $\zeta$, we re-render the Gaussian images and apply the repair model once again, replacing the previously repaired images for further supervision. This meticulous refinement process leads to high-quality outcomes in both foreground and background scenes.

By combining the strengths of different large model priors, LM-Gaussian can synthesize new views with competitive quality and superior details compared to state-of-the-art methods in sparse-view settings, particularly in 360-degree scenes. The contributions of our method can be summarized as follows:
\begin{itemize}
\item We propose a new method capable of generating high-quality novel views in a sparse-view setting with large model priors. Our method surpasses recent works in sparse-view settings, especially in large-scale 360-degree scenes.

\item We introduce a Background-Aware Depth-guided Initialization Module, capable of simultaneously reconstructing high-quality dense point clouds and camera poses for initialization.

\item We introduce a Multi-modal Regularized Gaussian Reconstruction Module that leverages regularization techniques from various domains to avoid overfitting issues.

\item We present an Iterative Gaussian Refinement Module which uses diffusion priors to recover scene details and achieve high-quality novel view synthesis results.
\end{itemize}

\section{Related work}
\boldparagraph{3D Representations for Novel-view synthesis} 
Novel view synthesis (NVS) involves rendering unseen viewpoints of a scene from a given set of images. One popular approach is Neural Radiance Fields (NeRF), which uses a Multilayer Perceptron (MLP) to represent 3D scenes and renders via volume rendering. Several works have aimed to enhance NeRF's performance by addressing aspects such as speed~\cite{muller2022instant,fridovich2022plenoxels,garbin2021fastnerf,li2023nerfacc}, quality~\cite{wang2022nerf, barron2021mip,wang2021neus,zhang2020nerf++}, and adapting it to novel tasks~\cite{weng2022humannerf, zhang2021nerfactor, su2021nerf, rudnev2022nerf}. While NeRF relies on a neural network to represent the radiance field, 3D Gaussian Splatting (3DGS)\cite{kerbl20233d} stands out by using an ensemble of anisotropic 3D Gaussians to represent the scene and employs differentiable splatting for rendering. This approach has shown remarkable success in efficiently and accurately reconstructing complex real-world scenes with superior quality. Recent works have further extended the capabilities of 3DGS to perform various downstream tasks, including text-to-3D generation\cite{tang2023dreamgaussian, yi2023gaussiandreamer,chen2024text, xu2024agg, yang2024gaussianobject, zhou2024dreamscene360, liang2024luciddreamer}, dynamic scene representation~\cite{luiten2023dynamic, wu20244d, ling2024align, zhou2024drivinggaussian, zeng2024stag4d, bae2024per,lu20243d, shao2024control4d, huang2024sc, lin2024gaussian, yu2024cogs,li2024spacetime,sun20243dgstream}, editing~\cite{chen2024gaussianeditor,wang2024gaussianeditor,ye2023gaussian,zhou2024feature}, compression~\cite{wang2024end, fan2023lightgaussian, lee2024compact, morgenstern2023compact}, SLAM~\cite{yan2024gs,keetha2024splatam,matsuki2024gaussian,huang2024photo}, animating humans~\cite{zielonka2023drivable, li2024animatable, moreau2024human, hu2024gauhuman,liu2024humangaussian,shao2024splattingavatar,abdal2024gaussian,qian2024gaussianavatars,zheng2024gps,hu2024gaussianavatar}, and other novel tasks~\cite{shi2024language, zhou2024feature, qin2024langsplat,huang20242d,xie2024physgaussian, zou2024triplane, guedon2024sugar,cheng2024gaussianpro,lin2024vastgaussian,liang2024gs,yan2024multi,jiang2024gaussianshader,lu2024scaffold,zhou2024drivinggaussian}.

% However, both NeRF and 3D Gaussian Splatting are primarily designed for dense view inputs, typically involving a larger number of views (e.g., 100-300 views). Additionally, these methods rely on preprocessing software, such as colmap, to compute camera extrinsics and intrinsics. The open challenge lies in reconstructing the entire scene using only a few unposed viewpoints, which remains an active area of research. 
% ，yang2024deformable

\boldparagraph{Sparse View Scene Reconstruction and Synthesis}
Sparse view reconstruction aims to reconstruct a scene using a limited number of input views.
Several studies~\cite{deng2022depth, xiong2023sparsegs, wang2023sparsenerf, li2024dngaussian} address this challenge by using depth regularization from monocular estimation models~\cite{ranftl2021vision} to prevent overfitting. Some approaches employ semantic~\cite{jain2021putting}, frequency~\cite{yang2023freenerf}, continuity~\cite{niemeyer2022regnerf} and correlation~\cite{zhang2025cor} regularization to guide training, though these are often effective in specific scenes and may lack detail. Stereo~\cite{chen2021mvsnerf, chen2024mvsplat, fan2024instantsplat} and image feature~\cite{yu2021pixelnerf, charatan2024pixelsplat} priors are also used to synthesize novel views across various scenes, aiding the training process. Gaussian-based methods~\cite{lu2024scaffold, mihajlovic2025splatfields} incorporate structured voxels and implicit latents to enhance view-adaptive performance.
 More recently, generative models are used in sparse view reconstruction. GeNVS~\cite{chan2023generative}, latentSplat~\cite{wewer2024latentsplat} and Sparsefusion~\cite{zhou2023sparsefusion} utilize rendering view-conditioned feature fields followed by 2D generative decoding to generate novel views. But these methods are category-specific and can't generalize well. DiffusionNerf~\cite{wynn2023diffusionerf} trains an RGBD denosing model to regularize geometry and color of a scene. SparseGS~\cite{xiong2023sparsegs} uses a SDS loss to distill information while ZeroNVS~\cite{sargent2023zeronvs} finetunes a view-conditioned diffusion model~\cite{liu2023zero} to enable single-image reconstruction. PERF~\cite{wang2024perf} use diffusion model to inpaint invisible areas. Reconfusion~\cite{wu2024reconfusion} and CAT3D~\cite{gao2024cat3d} train a view-conditioned image-to-image diffusion model to directly output novel view images. Although these methods have shown impressive results in sparse view reconstructions, they face challenges with high pretraining costs, limited input views and inconsistency with original input views.

\boldparagraph{Unposed Scene Reconstruction}
The methods mentioned above all rely on known camera poses, and Structure from Motion (SfM) algorithms often struggle to predict camera poses and point clouds with sparse inputs, mainly due to a lack of image correspondences. Therefore, removing camera parameter preprocessing is another active line of research. For instance, iNeRF~\cite{yen2021inerf} demonstrates that poses for new view images can be estimated using a reconstructed NeRF model. NeRFmm~\cite{wang2021nerf} concurrently optimizes camera intrinsics, extrinsics, and NeRF training. BARF~\cite{lin2021barf} introduces a coarse-to-fine positional encoding strategy for joint optimization of camera poses and NeRF. GARF~\cite{chng2022gaussian} illustrates that utilizing Gaussian-MLPs simplifies and enhances the accuracy of joint pose and scene optimization. Recent works like Nope-NeRF~\cite{bian2023nope}, LocalRF~\cite{meuleman2023progressively}, and CF-3DGS~\cite{yu2024mip} leverage depth information to constrain NeRF or 3DGS optimization. While demonstrating promising outcomes on forward-facing datasets such as LLFF~\cite{mildenhall2019local}, these methods encounter challenges when dealing with complex camera trajectories involving significant camera motion, such as 360-degree large-scale scenes.
\section{Preliminary}
\label{headings}
\subsection{3D Gaussian Splatting}
3D Gaussian Splatting (3D-GS) represents a 3D scene with a set of 3D Gaussians. Specifically, a Gaussian primitive can be defined by a center $\mu \in \mathbb{R}^3$, a scaling factor $s \in \mathbb{R}^3$, and a rotation quaternion $q \in \mathbb{R}^4$. Each 3D Gaussian is characterized by:
\begin{equation}
    G(x)=\frac{1}{\left(2\pi\right)^{3/2}\left|\Sigma\right|^{1/2}}e^{-\frac{1}{2}{\left(x - \mu\right)}^T\Sigma^{-1}\left(x - \mu\right)}
\label{eq:3dgs}
\end{equation}
where the covariance matrix $\Sigma$ can be derived from the scale $s$ and rotation $q$.

To render an image from a specified viewpoint, the color of each pixel $p$ is computed by blending $K$ ordered Gaussians $\left\{G_i \mid i=1, \cdots ,K\right\}$ that overlap with $p$ using the following blending equation:
\begin{equation}
\label{eq:colorblending}
c(p)=\sum_{i=1}^K c_i \alpha_i \prod_{j=1}^{i-1}\left(1-\alpha_j\right),
\end{equation}
where $\alpha_i$ is determined by evaluating a projected 2D Gaussian from $G_i$ at $p$ multiplied by a learned opacity of $G_i$, and $c_{i}$ represents the learnable color of $G_i$. The Gaussians covering $p$ are sorted based on their depths under the current viewpoint. Leveraging differentiable rendering techniques, all attributes of the Gaussians can be optimized end-to-end through training for view reconstruction.

% \textbf{Rasterizing Depth for Gaussians} 
\boldparagraph{Rasterizing Depth for Gaussians} 
Following the depth calculation approach introduced in RaDe-GS~\cite{zhang2024rade}, the center $\mu_i$ of a Gaussian $G_i$ is initially projected into the camera coordinate system as $\mu_i^{\prime}$. Upon obtaining the center value $(x_i', y_i', z_i')$ for each Gaussian, the depth $(x,y,z)$ of each pixel is computed as:
\begin{equation}
    d =  z_i' + \mathbf{p} \begin{pmatrix}\Delta x\\
    \Delta y \end{pmatrix}, 
    \mu_{i}^{\prime} = \left[\begin{array}{c}
x_{i}' \\
y_{i}' \\
z_{i}'
\end{array}\right]=\mathbf{W} \mu_{i}+\mathbf{t},
\label{eq:depth_render}
\end{equation}
where $z_i'$ represents the depth of the Gaussian center, $\Delta x = x_i' - x$ and $\Delta y = y_i' - y$ denote the relative pixel positions. The vector $\mathbf{p}$ is determined by the Gaussian parameters $[\mathbf{W}, \mathbf{t}] \in \mathbb{R}^{3 \times 4}$. 
 
% \textbf{Rasterizing Normal for Gaussians}
\boldparagraph{Rasterizing Normal for Gaussians}
In accordance with RaDe-GS, the normal direction of the projected Gaussian is aligned with the plane's normal. To compute the normal map, we transform the normal vector from the 'rayspace' to the 'camera space' as follows:
\begin{equation}
    \mathbf{n} = -\mathbf{J}^\top \left(
\frac{\mathbf{z_i'}}{\mathbf{z_i}} \mathbf{p} \quad 1 \right)^\top,
\end{equation}
where $\mathbf{J}$ represents the local affine matrix, and the vector $\mathbf{p}$ has been defined earlier. 

\subsection{Diffusion model}
In recent years, diffusion models have emerged as the state-of-the-art approach for image synthesis. These models are characterized by a predefined forward noising process $\{\mathbf{z_t}\}_{t=1}^{T}$ that progressively corrupts the data by introducing random noise $\epsilon$.
\begin{equation}
   z_t = \sqrt{\overline{\alpha}_t}z_0 + \sqrt{1-\overline{\alpha}_t}\epsilon, \epsilon \in  \mathbf{N(0, I)},
\label{eq:diffusion}
\end{equation}
where $t \in [1, T]$ denotes the time step and $\overline{\alpha}_t=\alpha_1\cdot\alpha_2...\alpha_t$ represents a decreasing sequence. These models can generate samples from the underlying data distribution given pure noise by training a neural network to learn a reversed denoising process. Having learned from hundreds of millions of images from the internet, diffusion priors exhibit a remarkable capacity to recover real-world details.

\begin{table}[!t]
\caption{\textbf{Symbol definition.} For clarity, we first definition symbols mentioned in this paper.}\label{tab:ablate}
\renewcommand{\arraystretch}{1.5} % 调整行高
\resizebox{1\linewidth}{!}{
\centering
\begin{tabular}{c|cccccccccc}
\hline
Symbol  & Definition   \\ \hline
% $\boldsymbol{I}$   & RGB input image\\ \hline
$\boldsymbol{I}_k$ & RGB input image of $k_{th}$ view     \\ \hline
% $\boldsymbol{P}$   & Point map: each pixel's value is the coordinates of a 3D point.   \\ \hline
$\boldsymbol{P}_{k}$ & 3D point map of $k_{th}$ view    \\ \hline

$\boldsymbol{\eta}_{k}$ & Confidence map  of $k_{th}$ view  \\  \hline
$\boldsymbol{\hat{D}_k}$, $\boldsymbol{\hat{N}_k}$  & Monocular estimated depth / normal map of $k_{th}$ view \\ \hline
$\boldsymbol{\bar{I}_k}$, $\boldsymbol{\bar{D}_k}$, $\boldsymbol{\bar{N}_k}$ & Gaussian-rendered RGB / depth / normal image in $k_{th}$ view.  \\ \hline
\end{tabular}}
\end{table}
\section{Method}

\begin{figure*}
\includegraphics[width=\textwidth]{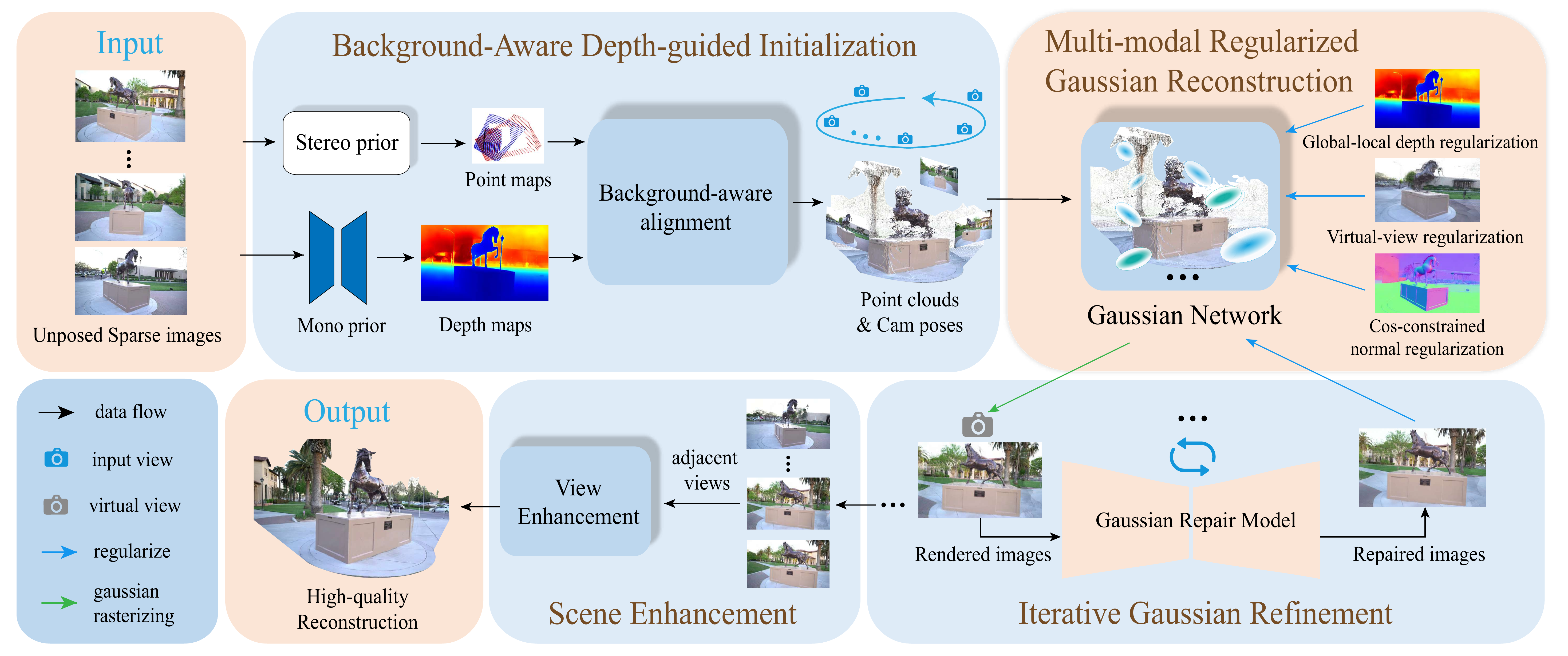}
\caption{\textbf{The Framework of LM-Gaussian.} Our method takes unposed sparse images as inputs. For example, we select 8 images from the Horse Scene to cover a 360-degree view. Initially, we utilize a Background-Aware Depth-guided Initialization Module to generate dense point clouds and camera poses (see Section \ref{sec:Initialization}). These variables act as the initialization for the Gaussian kernels. Subsequently, in the Multi-modal Regularized Gaussian Reconstruction Module (see Section \ref{sec:Coarse}), we collectively optimize the Gaussian network through depth, normal, and virtual-view regularizations. After this stage, we train a Gaussian Repair model capable of enhancing Gaussian-rendered new view images. These improved images serve as guides for the training network, iteratively restoring Gaussian details (see Section \ref{sec:Refine}). Finally, we employ a scene enhancement module to further enhance the rendered images for realistic visual effects (see Section \ref{sec:Enhance}). The image of the point maps is sourced from DUSt3R~\cite{wang2024dust3r}}. \label{fig1}
\end{figure*}

\subsection{Overview}
In this paper, we introduce a new method called LM-Gaussian, which aims to generate high-quality novel views of 360-degree scenes using a limited number of input images. Our approach integrates multiple large model priors and is composed of four key modules:
1) \textbf{Background-Aware Depth-guided Initialization}:  This module extends DUSt3R for camera pose estimation and detailed 3D point cloud creation. By integrating depth priors and point cleaning, we achieve a high-quality point cloud for Gaussian initialization (see Section \ref{sec:Initialization}).
2) \textbf{Multi-Modal Regularized Gaussian Reconstruction}: In addition to the photometric loss used in 3DGS, we incorporate depth, normal, and virtual-view constraints to regularize the optimization process (see Section \ref{sec:Coarse}).
3) \textbf{Iterative Gaussian Refinement}: We use image diffusion priors to enhance rendered images from 3DGS. These improved images further refine 3DGS optimization iteratively, incorporating diffusion model priors to boost detail and quality in novel view synthesis (see Section \ref{sec:Refine}).
4) \textbf{Scene Enhancement}: In addition to image diffusion priors, we apply video diffusion priors to further enhance the rendered images from 3DGS, enhancing the realism of visual effects (refer to Section \ref{sec:Enhance}).

\subsection{Background-Aware Depth-guided Initialization } 
\label{sec:Initialization}
Traditionally, 3DGS relies on point clouds and camera poses calculated through Structure from Motion (SfM) methods for initialization. However, SfM methods often encounter challenges in sparse view settings. To address this issue, we propose leveraging stereo priors~\cite{wang2024dust3r} as a solution. DUSt3R, an end-to-end dense stereo model, can take sparse views as input and produce dense point clouds along with camera poses. Nevertheless, the point clouds generated by DUSt3R are prone to issues such as floating objects, artifacts, and distortion, particularly in the background of the 3D scene.

To overcome these challenges, we introduce the Background-Aware Depth-guided Initialization module to generate dense and precise point clouds. This module incorporates four key techniques:
1) \textbf{Camera Pose Recovery}: Initially, sparse images are used to generate point clouds for each image using DUSt3R. Subsequently, the camera poses and point clouds are aligned into a globally consistent coordinate system.
2) \textbf{Depth-guided Optimization}: Depth-guided optimization is then employed to refine the aligned point cloud. In this step, a monocular estimation model is used as guidance for the optimization process.
3) \textbf{Point Cloud Cleaning}: Two strategies are implemented for point cloud cleaning: geometry-based cleaning and confidence-based cleaning. During optimization, after every $\xi$ iterations, a geometry-based cleaning step is executed to remove unreliable floaters. Following the optimization process, confidence-based cleaning is applied to distinguish between foreground and background, utilizing specific filtering techniques to preserve the final output point cloud. 
Next, we will provide detailed insights into the implementation of each component within this module.

\boldparagraph{Camera pose recovery} 
% To recover camera poses, a pose graph similar to DUSt3R is constructed to align all poses and point clouds into a unified coordinate system. With $K$ poses, all $\frac{K(K-1)}{2}$ edges are exhaustively sorted by confidence, and the top $K-1$ edges are selected to connect all poses. Subsequently, point maps, where each pixel's value corresponds to its 3D point's coordinates, are aligned by arranging a list of input images and calculating the scale, rotation, and translation between predicted and aligned point maps. After establishing the pose graph, camera poses and point clouds are aligned within a unified coordinate system.
We first use minimum spanning tree algorithm~\cite{kummerle2011g} to align all camera poses and point clouds into a unified coordinate system. An optimization scheme is then utilized to enhance the quality of the aligned point clouds. Initially, following the approach of DUSt3R, a point cloud projection loss $\mathcal{L}_{pc}$ is minimized. Consider the image pair  $\{\boldsymbol{I}_k, \boldsymbol{I}_l\}$ where $\boldsymbol{P}_{k}$ and $\boldsymbol{P}_{l}$ denote the point map in the  $k_{th}$ and $l_{th}$ camera's coordinate system. The objective is to evaluate the consistency of 3D points in the $k_{th}$ coordinate system with those in the $l_{th}$ coordinate system. The projection loss is computed by projecting the point map $\boldsymbol{P}_{l}$ to the $k_{th}$ coordinate system using a transformation matrix $\boldsymbol{T}_{k,l}$ that converts from the $l_{th}$ coordinate system to the $k_{th}$ coordinate system. The loss parameters include the transformation matrix $\boldsymbol{T}_{k,l}$, a scaling factor $\sigma_{k,l}$, and $\boldsymbol{P}_k$. This process is repeated for the remaining image pairs.
\begin{equation}
\mathcal{L}_{pc} = \sum_{k\in K} \sum_{l\in K\setminus \{k\}} \boldsymbol{\eta}_k\cdot\boldsymbol{\eta}_l \left\Vert \boldsymbol{P}_k - \sigma_{k,l} \boldsymbol{T}_{k,l} \boldsymbol{P}_{l} \right\Vert
\label{eq:global_align}
\end{equation}

The purpose of this loss function is to systematically pair each input image like $\boldsymbol{I}_k$ with all other images such as $\boldsymbol{I}_l$. For the image pair $\{\boldsymbol{I}_k, \boldsymbol{I}_l\}$, the loss function measures the disparity between the point map $\boldsymbol{P}_{k}$ in the $k_{th}$ coordinate system and the transformed point map $\sigma_{k,l}\boldsymbol{T}_{k,l}\boldsymbol{P}_{l}$. These comparisons are weighted by their respective confidence maps $\boldsymbol{\eta}_k$ and $\boldsymbol{\eta}_l$.

\boldparagraph{Depth-guided Optimization}
The optimization based solely on the projection loss may not be sufficient for reconstructing large-scale scenes, as it could lead to issues like floaters and scene distortion that can impact subsequent reconstructions. To tackle scene distortion, we integrate a robust depth prior to guide the optimization network. Recently many depth estimation models~\cite{yang2024depth}~\cite{hu2024metric3d}~\cite{fu2025geowizard} show great performance, here we use Marigold~\cite{ke2024repurposing} to provide insights into the scene's depth information.

The monocular depth estimation model significantly enhances depth perception across different scales. Its guidance is pivotal in mitigating distortion issues and improving overall scene depth perception.
Within the optimization network, we merge DUSt3R outputs with depth guidance by incorporating a point cloud projection loss, a multi-scale depth loss and a depth smoothness loss.

\begin{equation}
\mathcal{L}_{opt} = \mathcal{L}_{pc} + \alpha_d\mathcal{L}_{D} + \alpha_s\mathcal{L}_{smooth}
\end{equation}
where $\mathcal{L}_{opt}$ refers to the total optimization loss. $\alpha_d$ and $\alpha_s$ are loss weights of multi-scale depth loss $\mathcal{L}_{D}$ and depth smoothness loss $\mathcal{L}_{smooth}$. The smoothness loss encourages depth map smoothness by penalizing depth gradient changes, weighted by the image gradients and details of the multi-scale depth loss term $\mathcal{L}_{D}$ would be discussed later (see Sec~\ref{sec:Coarse}).

\begin{figure}
\includegraphics[width=0.48\textwidth]{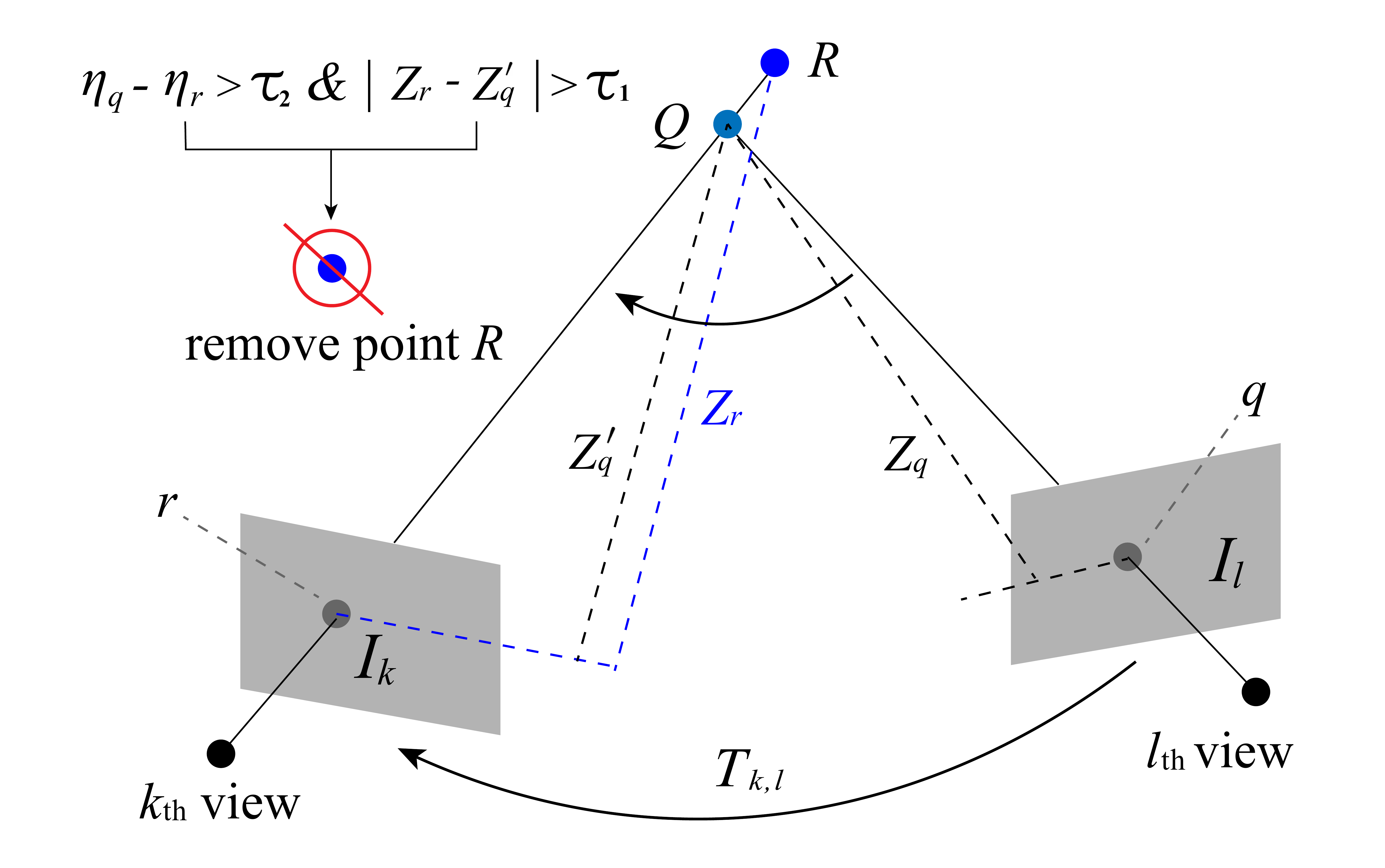}
\caption{\textbf{The depicted curve illustrates the point cleaning operation.} We project 3D point $Q$ from the $l_{th}$ coordinate system into the $k_{th}$ coordinate system. If the difference between the projected depth $Z_q'$ and the reference depth $Z_r$ exceeds a threshold $\tau_1$, and the confidence $\eta_q$ of point $Q$ exceeds the confidence $\eta_r$ of point $R$ by more than $\tau_2$, we classify point $R$ as artifacts and proceed to remove them. Otherwise we keep both two points.} \label{clean}
\end{figure}

\boldparagraph{Point cloud Cleaning}
\label{pointclean}
In order to eliminate floaters and artifacts, we implement two strategies for cleaning the point cloud: geometry-based cleaning and confidence-based cleaning.

In \textbf{geometry-based cleaning}, we adopt an iterative approach to remove unreliable points during the depth optimization process. For a set of $K$ input images, as illustrated in Figure \ref{clean}, the method involves systematically pairing image $\boldsymbol{I}_k$ with all other $K-1$ images within a single iteration. For the image pair ${\boldsymbol{I}_k, \boldsymbol{I}_l}$, a pixel $q$ in $\boldsymbol{I}_l$ corresponds to a 3D point $Q$ in the scene, represented as $(X_q, Y_q, Z_q)$ in the $l_{th}$ coordinate system. This point can be translated into the $k_{th}$ coordinate system using the transformation matrix $\boldsymbol{T}_{k,l}$. The projected point intersects with $\boldsymbol{I}_k$ at pixel $r$, and its depth in the $k_{th}$ coordinate system is denoted as $Z_q'$. Conversely, pixel $r$ also corresponds to another 3D point $R$, denoted as $(X_r, Y_r, Z_r)$ in the $k_{th}$ coordinate system.

To tackle the floating issue, if we detect that the difference between the projected depth $Z_q'$ and the depth $Z_r$ of point $R$ exceeds a threshold $\tau_1$, and the confidence $\eta_q$ of point $Q$ exceeds the confidence $\eta_r$ of point $R$ by more than $\tau_2$, we label point $R$ as unreliable. As a result, we remove this point from the set of 3D points.
\begin{equation}
\left|Z_r - Z_q'\right| > \tau_1 \text{ and} \quad \eta_q - \eta_r > \tau_2 \Rightarrow \text{Exclude point } R
\end{equation}
The cleaning operation of the point clouds is executed once every $\xi$ iterations, with $\tau_1$ and $\tau_2$ serving as hyperparameters.

\begin{figure}
\includegraphics[width=0.48\textwidth]{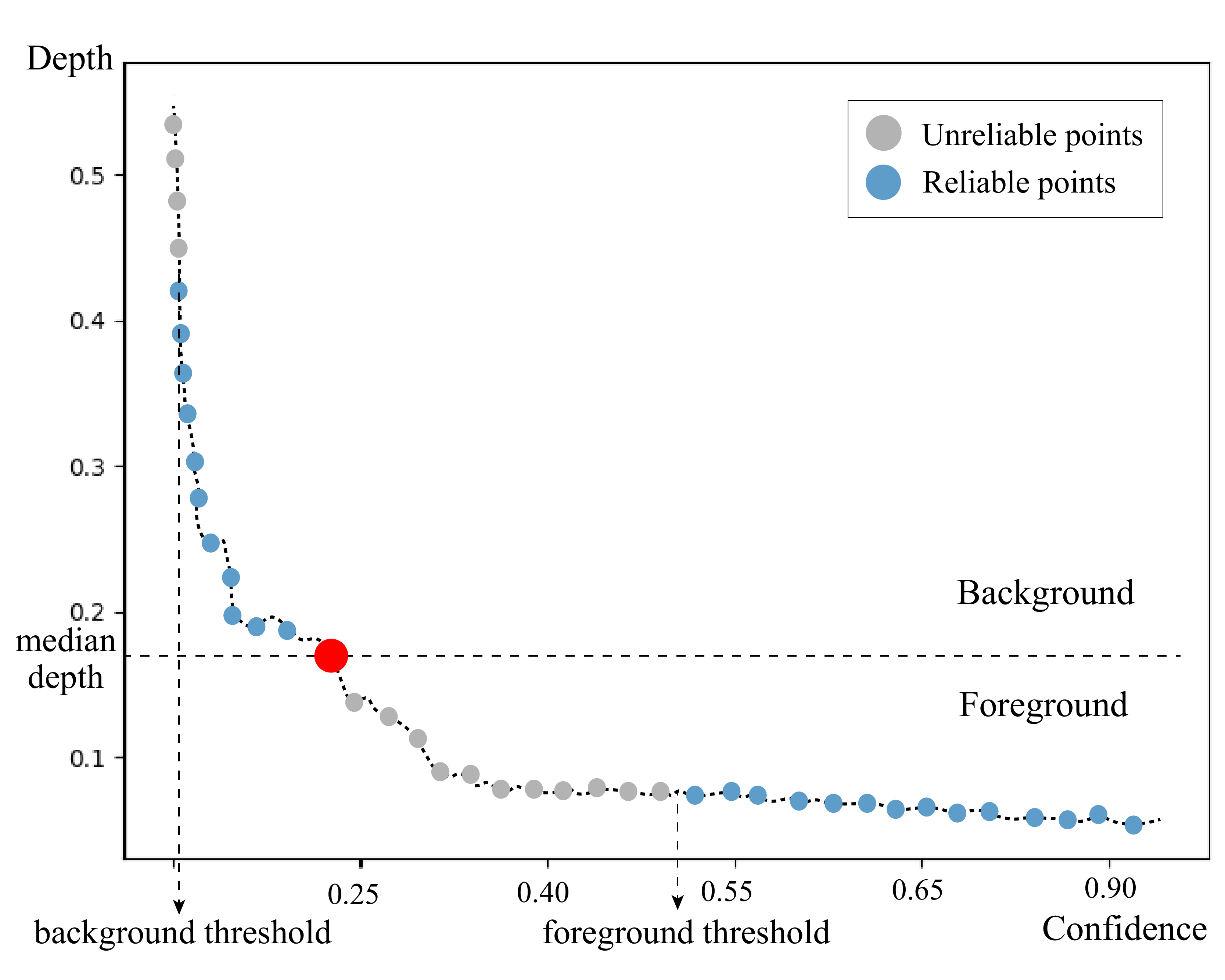}

\caption{The diagram depicted showcases the relationship between depth values and confidence of the 3D points. Points with a large distance exhibit decreased reliability, due to the bias of DUSt3R. To address this problem, we first divide the points into foreground and background parts based on median depth. Instead of using a single confidence threshold for the whole scene, we use two separate confidence thresholds to process the foreground and background parts individually.} \label{separate}
\end{figure}

In addition to the geometry-based cleaning process, we also implement a \textbf{confidence-based cleaning} step post-optimization. Each point within the point clouds is assigned a confidence value. The original DUSt3R method applies a basic confidence threshold to filter out points with confidence below a certain level. However, due to the distance bias, this approach may inadvertently exclude many background elements. To tackle this challenge, as depicted in Figure \ref{separate}, we differentiate between foreground and background regions by arranging the depths of all points and selecting the median depth as the separation boundary. Points in the foreground, typically observed in multiple-view images, tend to have higher confidence levels. Consequently, we establish a high-confidence threshold for foreground objects. On the contrary, the background area, often captured from a distance and present in only a few images, tends to exhibit lower confidence levels. Hence, we adopt a more lenient strategy for this region, employing a lower confidence threshold for point cleaning.

\subsection{Multi-modal Regularized Gaussian Reconstruction}
\label{sec:Coarse}
Dense point clouds and camera poses are acquired through Background-Aware Depth-guided Initialization. These variables serve as the initialization of Gaussian kernels. Vanilla 3DGS methods utilize photo-metric loss functions such as $\mathcal{L}_1$ and $\mathcal{L}_{SSIM}$ to optimize 3DGS kernels and enable them to capture the underlying scene geometry.
However, challenges arise in scenarios with extremely sparse input images. Due to the inherent biases of the Gaussian representation, the Gaussian kernels are prone to overfitting on the training views and cause degradation on unseen perspectives. To mitigate this issue, we enhance the Gaussian optimization process by integrating photo-metric loss, multi-scale depth loss, cosine-constrained normal loss, and norm-weighted virtual-view loss.

\boldparagraph{Photo-metric Loss} In line with vanilla 3DGS, we initially compute the photo-metric loss between the input RGB images and Gaussian-rendered images. The photo-metric loss function combines $\mathcal{L}_1$ with an SSIM term $\mathcal{L}_{SSIM}$. 
\begin{equation}
\label{finalloss}
\mathcal{L}_{pho} = (1-\lambda) \mathcal{L}_{1}+\lambda \mathcal{L}_{SSIM}
\end{equation}
where $\lambda$ represents a hyperparameter, and $\mathcal{L}_{pho}$ denotes the photo-metric loss.

\boldparagraph{Multi-scale Depth Regularization} 
To mitigate overfitting, depth regularization is incorporated into the Gaussian scene. Similar to NeRDi~\cite{deng2023nerdi}, we employ the Pearson Correlation Coefficient (PCC) \cite{cohen2009pearson} to assess the similarity between depth maps. To capture both global and local structures, we compute similarity both for entire images and for individual image patches.

The Pearson Correlation Coefficient is a fundamental statistical correlation coefficient that quantifies the linear correlation between two data sets. Essentially, it assesses the resemblance between two distinct distributions $X$ and $Y$.
\begin{equation}
\label{pccloss}
\operatorname{PCC}(X,Y) = \frac{E[XY] - E[X]E[Y]}{\sqrt{E[Y^2] - E[Y]^2} \sqrt{E[X^2] - E[X]^2}}
\end{equation}
where $E$ represents the mathematical expectation.

Similar to the Initialization Module, we initially employ the monocular estimation model Marigold~\cite{ke2024repurposing} to predict depth images $\{\hat{\boldsymbol{D}}_k\}_{k=0}^{K-1}$ from sparse input images. Then PCC is used to assess the similarity between Gaussian-rendered depth maps $\boldsymbol{D}_{gs}$ and estimated depth maps $\boldsymbol{D}_{mo}$ in a global level.
\begin{gather}
\mathcal{L}_{global} =  1 -  \operatorname{PCC}(\boldsymbol{D}_{gs}, \boldsymbol{D}_{mo}) 
\end{gather}
Inspired by previous works~\cite{li2024dngaussian}~\cite{xiong2023sparsegs}, to enhance the capture of local structures, we divide depth images into small patches and compare the correlation among these depth patches. During each iteration, we randomly select $F$ non-overlapping patches to evaluate the depth correlation loss, defined as:
\begin{gather}
\mathcal{L}_{local} = \frac{1}{F} \sum^{F-1}_{f=0} 1 -  \operatorname{PCC}(\boldsymbol{\bar{\Gamma}}_f, \boldsymbol{\hat{\Gamma}}_f) 
\end{gather} where $\boldsymbol{\bar{\Gamma}}_f$ denotes the $f_{th}$ patch of Gaussian-rendered depth maps and $\boldsymbol{\hat{\Gamma}}_f$ denotes the $f_{th}$ patch of depth maps predicted by monocular estimation model. 
\begin{gather}
\mathcal{L}_{depth} = \mathcal{L}_{global} + \mathcal{L}_{local}
\end{gather} 
Here the $\mathcal{L}_{depth}$ means the multi-scale depth loss. Intuitively, this loss works to align depth maps of the Gaussian representation with the depth map of monocular prediction, mitigating issues related to inconsistent scale and shift.

\boldparagraph{Cosine-constrained Normal Regularization} While depth provides distance information within the scene, normals are also essential for shaping surfaces and ensuring smoothness. Therefore, we introduce a normal-prior regularization to constrain the training process.

% The Marigold Normal Model, a diffusion-based model specializing in surface normals, is employed. Leveraging the rich visual knowledge embedded in Stable Diffusion, the Marigold Normal Model exhibits remarkable proficiency in normal prediction tasks. In our framework, this model is utilized for normal map prediction.

Similar to MonoSDF ~\cite{yu2022monosdf}, we utilize cosine similarity to quantify the variance between the predicted normal maps got from normal prior~\cite{ke2023repurposing} and the normal maps rendered using Gaussian representations.
\begin{gather}
\mathcal{L}_{normal} = \frac{1}{K} \sum^{K-1}_{k=0} 1 - \operatorname{COS}(\boldsymbol{\bar{N}}_k, \boldsymbol{\hat{N}}_k) 
\end{gather}
where $\boldsymbol{\bar{N}}_k \in \mathbb{R}^{H\times W\times 3}$ represents the Gaussian-rendered normal maps, and $\boldsymbol{\hat{N}}_k$ signifies the normal maps predicted by the monocular estimation model. The function $\operatorname{COS}()$ denotes the cosine similarity function.

\begin{figure*}
\includegraphics[width=\textwidth]{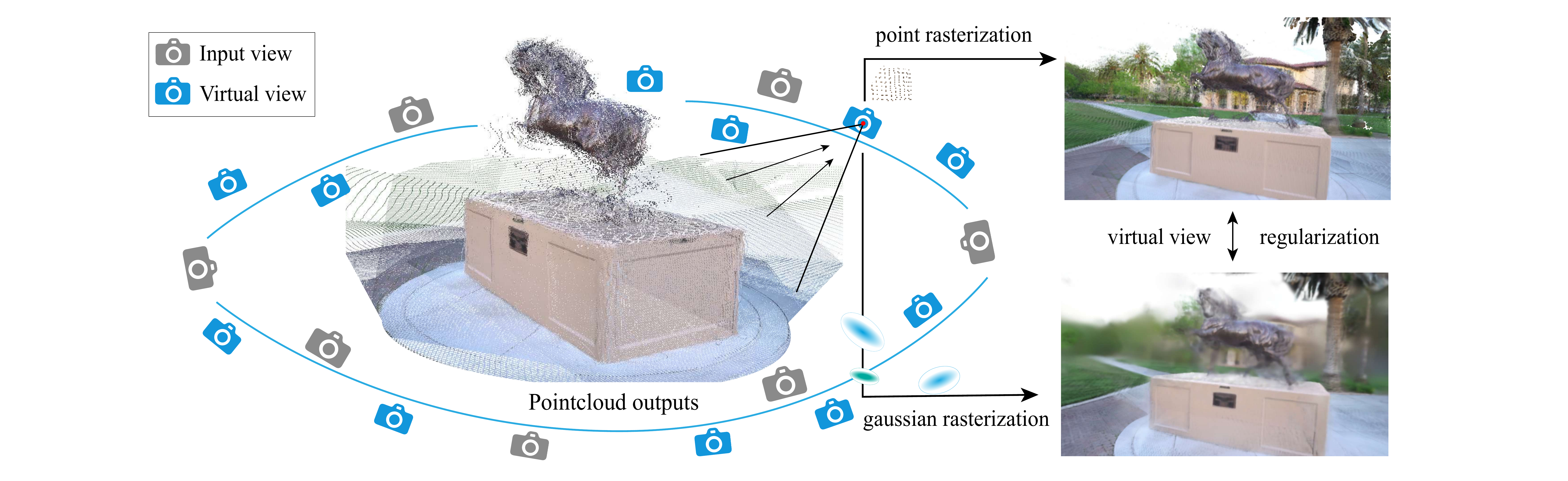}
\caption{\textbf{Visual Process of Weighted Virtual View Regularization.} For each virtual view, we employ two distinct methods for image rasterization. First, we utilize Gaussian kernels to produce a Gaussian-rendered image. Then, we apply a weighted blending algorithm to create a point-rendered image. We enforce consistency between these two images.} \label{pointrender}
\end{figure*}

\boldparagraph{Weighted Virtual-view Regularization} 
In cases where the training views are sparse, the Gaussian scene may deteriorate when presented with new views due to the lack of supervision. Hence, we introduce a virtual-view regularization strategy to preserve the original point cloud information throughout the optimization process.

As illustrated in Figure \ref{pointrender}, we randomly sample $K_v$ virtual views in 3D space. For each virtual camera, we project the point clouds onto the 2D plane of the view. A weighted blending algorithm is employed to render the 3D points into RGB point-rendered images. These point-rendered images serve as guidance for the Gaussian optimization process.

\begin{figure}
\includegraphics[width=0.5\textwidth]{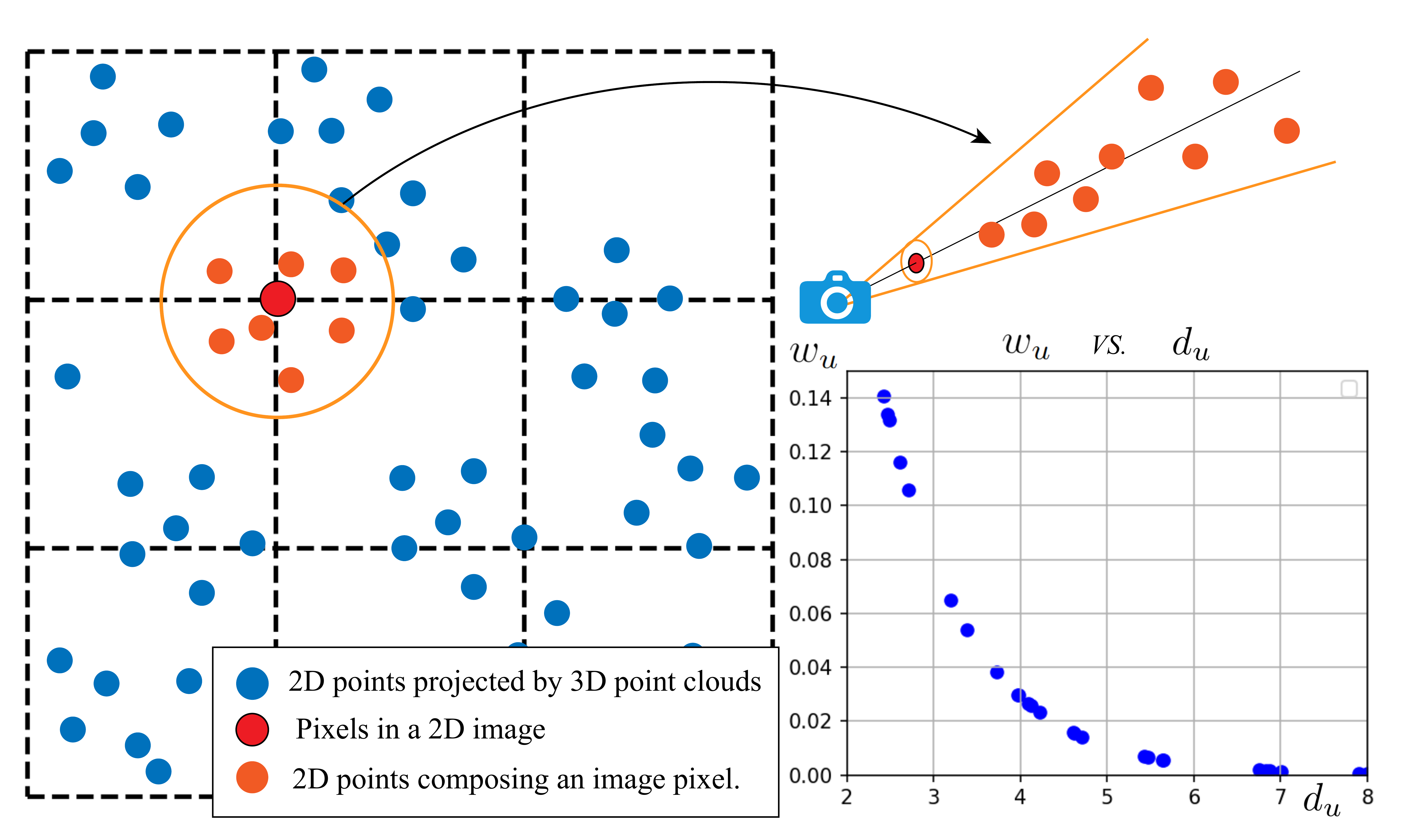}
\caption{\textbf{Illustration of Weighted Blending for an Image Pixel.} The blue point represents the 2D projection of a point from the 3D point clouds. The red point corresponds to a pixel on the RGB image. The orange points indicate the selected points that contribute to the final color of the red pixel. The scatter plot demonstrates the diverse weights assigned to 3D points at different depths, with $U$ fixed at 30. Relevant implementation is based on pytorch3d~\cite{ravi2020pytorch3d}} \label{pointraster}
\end{figure}

When creating a point-rendered RGB image from a virtual view, the color of each pixel $i$ is determined by the $U$ nearest projected 3D points. As shown in Figure \ref{pointraster}, these points are selected based on their proximity to pixel $i$ within a radius $\pi$, where $\pi$ is defined as one-third of the pixel width. Subsequently, these selected points are arranged in order of their distances $\{d_u\}_{u=0}^{U-1}$ from the viewpoint. Weights are then allocated to these ordered points according to their distances, with closer 3D points to the virtual viewpoint receiving higher weights.
% \[
% c(i) =
% \begin{cases} 
% \sum_{u=0}^{U-1} c_u w_u,\quad
% w_u = \frac{e^{-d_u}}{\sum_{u=0}^{U-1} e^{-d_u}} & \text{if valid points} \\
% c_{bg} & \text{otherwise}
% \end{cases}
% \]
\begin{equation}
c(i) =
\begin{cases} 
\sum_{u=0}^{U-1} c_u w_u,\quad
w_u = \frac{e^{-d_u}}{\sum_{u=0}^{U-1} e^{-d_u}} & \text{if valid points} \\
c_{bg} & \text{otherwise}
\end{cases}
\end{equation}
Here, $c(i)$ represents the color of pixel $i$ after point rasterization. $c_u$ denotes the color of the $u_{th}$ 3D point, and $w_u$ is its corresponding weight. In cases where a pixel has no valid point projection (i.e., $U=0$), we assign the pixel the predefined background color $c_{bg}$, which, in this instance, is white.

By employing the norm-weighted blending algorithm, we obtain $K_v$ point-rendered RGB images denoted as $\{\boldsymbol{I}_k^{pr}\}_{k=0}^{K_v-1}$. These images are subsequently utilized to regulate Gaussian kernels, thereby imposing constraints on optimization and preventing overfitting. The virtual-view loss function at this stage is presented below.
\begin{equation}
\begin{split}
 \mathcal{L}_{vir} =(1-\lambda) \mathcal{L}_1(\boldsymbol{\bar{I}}_k,\boldsymbol{I}_k^{pr}) + \\\lambda \mathcal{L}_{SSIM}(\boldsymbol{\bar{I}}_k,\boldsymbol{I}_k^{pr}),k \in K_v
 \end{split}
\end{equation}
where $\lambda$, $\mathcal{L}_1$, $\mathcal{L}_{SSIM}$ are the same as original 3d Gaussian splatting and $\boldsymbol{\bar{I}}_k$ is the Gaussian-rendered RGB image from $k_{th}$ view.

\boldparagraph{Multi-modal Joint Optimization} Throughout the Multi-modal Regularized Gaussian Reconstruction phase, in addition to the photo-metric loss, Multi-scale Depth Regularization, Cosine-constrained Normal Regularization, and Norm-weighted Virtual-view Regularization are incorporated to steer the training process. These methodologies are pivotal in alleviating overfitting and upholding the output quality.
\begin{equation}
\label{finalloss}
\mathcal{L}_{multi} = \mathcal{L}_{pho}+\beta_{vir} \mathcal{L}_{vir} + \beta_{dep} \mathcal{L}_{\text {depth}} + \beta_{nor} \mathcal{L}_{\text {normal}}
\end{equation}
where $\mathcal{L}_{multi}$ represents the loss function utilized in the Multi-modal Regularized Gaussian Reconstruction. The weights $\beta_{vir}, \beta_{dep}, \beta_{nor}$ serve as hyperparameters to regulate their impact, with further elaboration provided in the Experiment Section.

\subsection{Iterative Gaussian Refinement}
\label{sec:Refine}
% After integrating the multi-modal regularized Gaussian reconstruction in Section \ref{sec:Coarse}, we achieve an initial outcome. While the aforementioned module effectively tackles overfitting issues, the resultant scene may still lack intricate details due to the limited input data.
During this phase, we implement an iterative optimization approach to progressively enhance scene details. Initially, we uniformly enhance the Gaussian-rendered images from virtual viewpoints using a Gaussian repair model. This model refines blurry Gaussian-rendered images into sharp, realistic representations. Following this enhancement, these refined images act as supplementary guidance, facilitating the optimization of Gaussian kernels in conjunction with depth and normal regularization terms. After $\zeta$ optimization steps, we re-render the Gaussian images and subject them to the repair model once more, replacing the previously refined images for another iteration of supervision.

\begin{figure*}
\includegraphics[width=\textwidth]{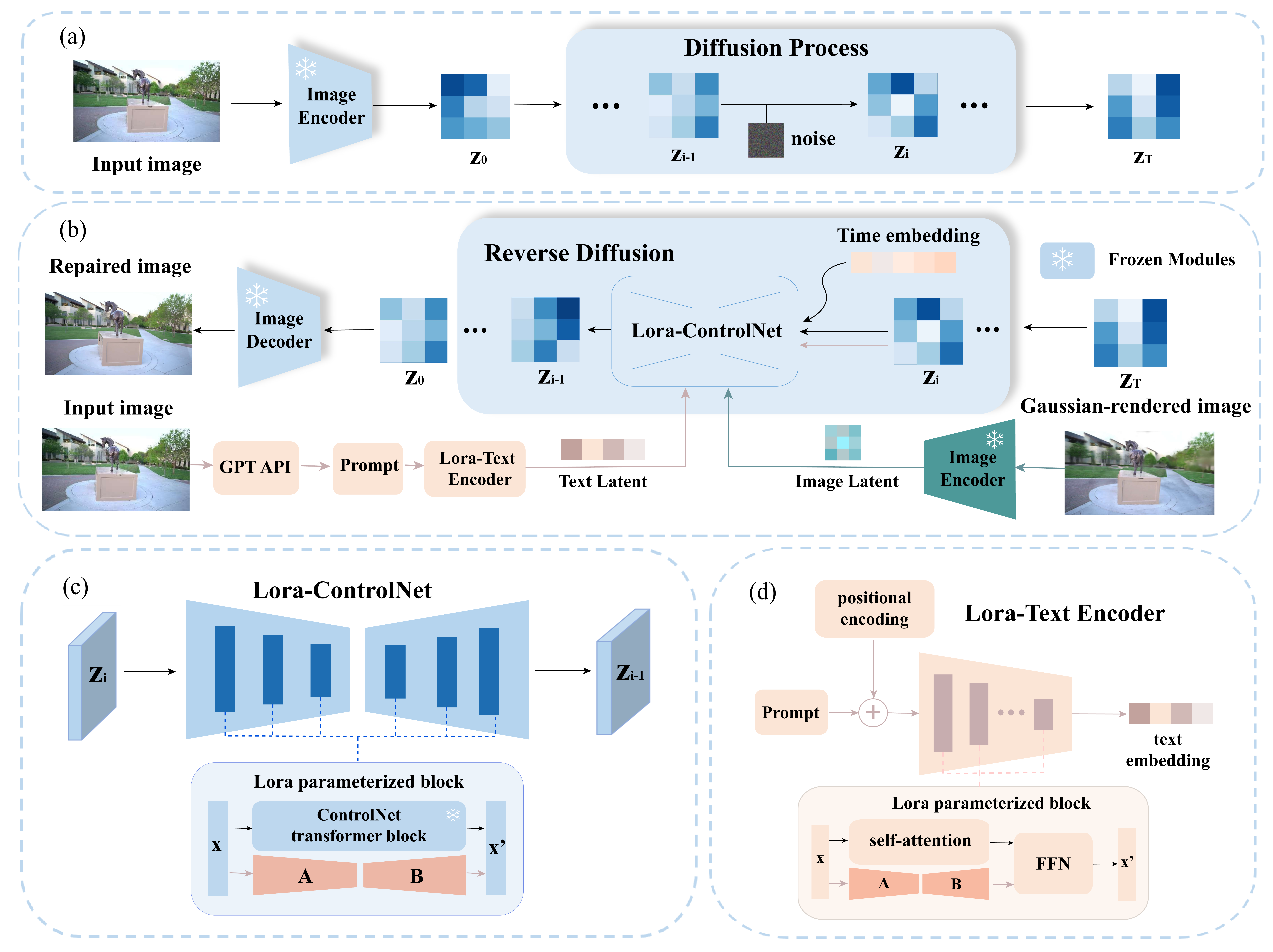}
\caption{\textbf{Architecture of Our Gaussian Repair Model.} The architectural layout of our Gaussian Repair Model entails selectively freezing all parameters except for the Lora weight within the network to fine-tune a ControlNet. This process imbues scene characteristics into the model, augmenting its ability to rectify blurry scene images. Precisely, we integrate Lora parameters into the text condition encoder and ControlNet's UNet. With a Lora rank designated as 16, this integration takes place in each transformer block, linear layer, and convolutional layer. By training on image pairs from the coarse stage, our model excels in generating detailed real-world images from blurry Gaussian-rendered counterparts, providing guidance for subsequent optimization.} \label{repairmodel}
\end{figure*}

\subsubsection{\textbf{Iterative Gaussian Optimization}}
\label{iterative}
% During the Gaussian refinement phase, we employ an iterative optimization approach to restore intricate scene details. 
Initially, we rasterize $K_v$ virtual view images from the current Gaussian scene. Subsequently, the Gaussian Repair Model is utilized to uniformly enhance these images, resulting in a set of repaired images denoted as $\{\boldsymbol{I}_k^{repair}\}_{k=0}^{K_v-1}$.
To maintain scene coherence and reduce potential conflicts, we set the denoise strength to a low value and gradually reintroduce limited details to the Gaussian-rendered images during each repair process. These repaired virtual-view images, along with monocular depth and normal maps from training views as outlined in Section \ref{sec:Coarse}, are then employed to regulate the Gaussian optimization. The overall optimization loss in the Gaussian refinement stage, denoted as $\mathcal{L}_{refine}$, is determined by:
\begin{equation}
\label{finalloss}
\mathcal{L}_{refine} = \mathcal{L}_{pho}+\beta_{rep} \mathcal{L}_{rep} + \beta_{dep} \mathcal{L}_{depth} + \beta_{nor} \mathcal{L}_{normal}
\end{equation}
Here, $\mathcal{L}_{depth}$ and $\mathcal{L}_{normal}$ correspond to the Multi-modal Regularized Gaussian Reconstruction. $\beta_{rep}$ signifies the weight of the repair loss. The repair loss $\mathcal{L}_{rep}$ is defined as follows:
\begin{equation}
\begin{split}
 \mathcal{L}_{rep} =(1-\lambda) \mathcal{L}_1(\boldsymbol{\bar{I}}_k,\boldsymbol{I}_k^{repair}) + \\\lambda \mathcal{L}_{SSIM}(\boldsymbol{\bar{I}}_k, \boldsymbol{I}_k^{repair}),k \in K_v
 \end{split}
\end{equation}
In this formulation, we leverage the photo-metric loss between the repaired images $\{\boldsymbol{I}_k^{repair}\}_{k=0}^{K_v-1}$and the Gaussian-rendered images $\{\boldsymbol{\bar{I}}_k\}_{k=0}^{K_v-1}$ in one loop, utilizing the repaired images as a guiding reference. The parameters $\lambda$, $\mathcal{L}_1$, and $\mathcal{L}_{SSIM}$ remain consistent with the original 3D Gaussian splatting methodology. The above operations will be repeated.

% After completing $\zeta$ iterations of optimization, we proceed with the Gaussian repair process by re-rendering images from virtual viewpoints and utilizing the Gaussian Repair Model to obtain corresponding repaired images. These updated images are reintegrated as regularization inputs for Gaussian optimization. 
Through this iterative optimization strategy, the newly generated images gradually enhance in sharpness without being affected by blurring caused by view disparities. The optimization process persists until it is ascertained that the diffusion process no longer produces satisfactory outcomes, as evidenced by deviations from the initial scene or inconsistencies across various viewpoints.

\subsubsection{\textbf{Gaussian Repair Model}} In this section, we present the Gaussian repair model utilized earlier. Its primary objective is to enhance blurry Gaussian-rendered images into sharp, realistic images while preserving the style and content of the original image.
 
\boldparagraph{Model Architecture} The architecture of the Gaussian Repair Model is illustrated in Figure \ref{repairmodel}. It takes Gaussian-rendered images and real-world input images as inputs. In Figure \ref{repairmodel}(b), the Gaussian-rendered image $\boldsymbol{\bar{I}}$ undergoes image encoding to extract latent image features. The real-world input images are processed through a GPT4v~\cite{achiam2023gpt} API for a description prompt $\sigma$, then encoded to obtain text latent features. These image\new{s} and text latent features act as conditions for a ControlNet~\cite{zhang2023adding} to predict noise $\epsilon_\theta$ and progressively remove noise from the Gaussian-rendered image.  Inspired by GaussianObject~\cite{yang2024gaussianobject}, the model is a controlnet finetuned by injecting lora into its layer and it can produce the repaired Gaussian-rendered image. Figure \ref{repairmodel}(c) provides insight into the Lora-ControlNet, where Lora~\cite{hu2021lora} weights are integrated into each transformer layer of the ControlNet's UNet. We maintain the original parameters of the transformer blocks constant and solely train the low-rank compositions $A, B$, where $A \in \mathbb{R}^{d \times r}, B \in \mathbb{R}^{r \times k}$, with a rank $r \ll \min(d, k)$. Concerning the text encoder, as depicted in Figure \ref{repairmodel}(d), Lora weights are integrated into each self-attention layer of the encoder. The input of the Lora-Text Encoder is the scene prompt, and the output is the text embedding.

% \boldparagraph{Foundational Model} Given the similarity between our Gaussian repair task and super-resolution tasks, as well as the existence of specialized ControlNet~\cite{zhang2023adding} weights for image super-resolution tasks in the open-source community, we choose ControlNet as our foundational model.

\boldparagraph{Training process} In this section, we will explore the training process of the Gaussian repair model. Initially, for data preparation, we collect image pairs from Section \ref{sec:Coarse}, where input images within each scene act as reference images. For each training perspective, we randomly select $\omega$ Gaussian-rendered images from different optimization timesteps and pair them with input images to create training pairs. Subsequently, these image pairs are utilized to train our Gaussian Repair Model. As shown in Figure \ref{repairmodel}(a), the input image $\boldsymbol{I}$ undergoes a forward diffusion process. Specifically, the image is input into an image encoder to extract the latent representation $z_1$. This latent representation then undergoes a diffusion process where noise $\epsilon$ is gradually introduced over $T$ steps. After obtaining latent $z_T$, a reverse diffusion process is initiated, as illustrated in Figure \ref{repairmodel}(b), where a Lora-UNet and a Lora-ControlNet are employed to predict noise $\epsilon_\theta$ at each step. This predicted noise, combined with the noise introduced during the diffusion process, contributes to calculating the loss function, aiding in the training process. The loss function for any input image can be defined as follows.
\begin{equation}
\mathcal{L}_{Control} = E_{\boldsymbol{I},t,\boldsymbol{\bar{I}},\epsilon\in \mathcal{N}(0, 1)}[\lVert(\epsilon_\theta(\boldsymbol{I}, t, \boldsymbol{\bar{I}}, \sigma)-\epsilon)\rVert_2^2]
\end{equation}
Here, $\sigma$ refers to the text prompt of the reconstructed scene. $\boldsymbol{I}$ represents the actual input image, and $\boldsymbol{\bar{I}}$ is the Gaussian-rendered image obtained from Section \ref{sec:Coarse}. $E_{\boldsymbol{I},t,\boldsymbol{\bar{I}},\epsilon\in \mathcal{N}(0, 1)}$ denotes the expectation over the input image $\boldsymbol{I}$, the time step $t$, the text prompt $\sigma$, the condition image $\boldsymbol{\bar{I}}$, and the noise $\epsilon$ drawn from a normal distribution with mean 0 and standard deviation 1. $\epsilon_\theta$ indicates the predicted noise.

% \boldparagraph{Sampling process} The sampling process mirrors the reverse diffusion process depicted in Figure \ref{repairmodel}(b). The difference is that the latent variable $z_T$ is initialized as random noise. The conditional image is a relatively blurry Gaussian-rendered image from any viewpoint and the text prompt is a description of the reconstructed scene got from GPT. More details please refer the supplement materials.
% Through the Image encoder and Lora-text encoder, we obtain text and image latents. Subsequently, Lora-UNet and Lora-ControlNet are leveraged to denoise the Gaussian-rendered image. By adjusting the denoising strength, the Gaussian Repair Model can produce diverse output images. The denoising strength serves as a hyper parameter, which will be further discussed in \ref{iterative}.

\subsection{Scene Enhancement}
\label{sec:Enhance}

Given the sparse input images and the restricted training perspectives, it is expected that rendered images from adjacent new viewpoints may display discrepancies. In order to ensure high-quality and consistent rendering along a specified camera path, we propose a View Enhancement module, which utilizes video diffusion priors to improve the coherence of rendered images.

% \boldparagraph{Consistency Enhancement} 
This module concentrates on enhancing the visual consistency of rendered images without delving into Gaussian kernel refinement. Initially, multiple images are rendered along a predetermined camera trajectory and grouped for processing. Subsequently, a video diffusion UNet is employed to denoise these images to generate enhanced images.
In the video diffusion model, DDIM inversion~\cite{song2020denoising} is utilized to map Gaussian-rendered images back to the latent space, the formulation can be expressed as:
\begin{equation}
z_{t+1}=\frac{\overline{\alpha}_{t+1}}{\overline{\alpha}_t}z_t+(\frac{1}{\overline{\alpha}_{t+1}}-1-\frac{1}{\overline{\alpha}_t}-1)\epsilon_{\theta}(z_t,t,\sigma),
\end{equation}
where $t\in[1,T]$ is the time step and $\overline{\alpha}_t$ denotes a decreasing sequence that guides the diffusion process. $\sigma$ serves as an intermediate representation that encapsulates the textual condition.

The rationale behind mapping Gaussian-rendered images to a latent space is to leverage the continuous nature of the latent space, preserving relationships between different views. By denoising images collectively in the latent space, the aim is to enhance visual quality without sacrificing spatial consistency.
% This process helps in maintaining relationships between images and forms a basis for subsequent denoising steps. 
% Let $z_T$ represent the noise tensor, where $z_T$ follows a Gaussian distribution with zero mean and unit variance. 
% The DDIM~\cite{couairon2022diffedit} utilizes a pretrained network $\epsilon_{\theta}$ to execute T denoising diffusion steps. Each iteration is crafted to estimate the intrinsic noise and then reconstruct a less noisy version of the tensor $z_{t-1}$ from its noisy counterpart $z_t$.
In the scene enhancement model, we utilized Zeroscope-XL~\cite{huggingface2023zeroscope} as the video-diffusion prior and set the denoising strength to 0.1.

\section{Experiments}

\begin{figure*}
\includegraphics[width=\textwidth]{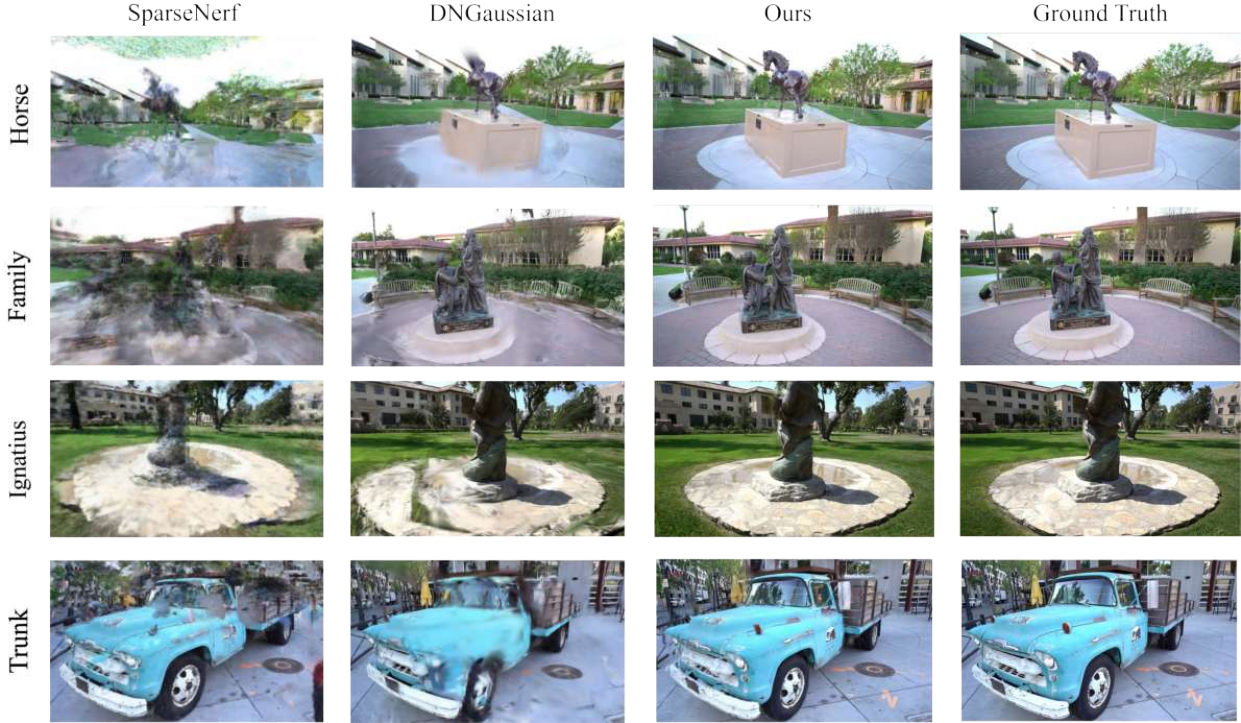}
\caption{\textbf{Qualitative comparison on Tanks and Temples Dataset with 16 input views.} Our approach consistently fairs better in recovering image structure from foggy geometry, where baselines typically struggle
with floaters and artifacts.} \label{tntresult}
\end{figure*}

\begin{table*}[thbp]
\newcommand{\xmark}{\ding{55}}
\centering
\caption{\new{Quantitative Comparisons with varying input views. Best results are highlighted as \colorbox{top1}{1st}, \colorbox{top2}{2nd} and \colorbox{top3}{3rd}.}}
\label{tab: finalresult}
\resizebox{\textwidth}{!}{
\begin{tabular}{c|l|ccc|ccc|ccc}
\toprule    
\multirow{2}{*}{} & \multirow{2}{*}{Method} & \multicolumn{3}{c|}{\new{4 views}} & \multicolumn{3}{c|}{8 views} & \multicolumn{3}{c}{\new{16 views}} \\
& & \new{SSIM$\uparrow$ }            & \new{PSNR$\uparrow$}  & \new{LPIPS$\downarrow$ }
& SSIM$\uparrow$             & PSNR$\uparrow$   & LPIPS$\downarrow$
&\new{SSIM$\uparrow$ }            & \new{PSNR$\uparrow$}  & \new{LPIPS$\downarrow$ } \\
\midrule 

\multirow{6}{*}[-1em]{\rotatebox[origin=c]{90}{Tanks\&Temples}} 
& FreeNeRF~\cite{yang2023freenerf}   & 0.2525   & 10.29  & 0.6025              & 0.2800  & 11.24  & 0.5400         & 0.3925 & 15.66 & 0.4375   \\
& SparseNerf~\cite{wang2023sparsenerf}    & 0.2625 & 10.35  & 0.6600              & 0.3000  & 11.45  & 0.5700        & 0.4600 & 16.20 & 0.4375 \\ 
& DNGaussian~\cite{li2024dngaussian} & 0.3025   & 11.59  & 0.6375              & 0.3200   & 12.67  & 0.5900        & 0.5025 & 16.69 & 0.4475   \\
& \new{Scaffold-GS}~\cite{lu2024scaffold}  & 0.3275   & 11.13  & 0.5600       & 0.4900   & 13.93  & 0.4675        & 0.5625 & 18.10 & 0.3600  \\
& \new{Splatfield}~\cite{mihajlovic2025splatfields}  & 0.3250 & 10.79  & 0.5975      & \cellcolor{top2}{0.5725}   & 14.17  & 0.5150        & 0.5725 & \cellcolor{top3}{18.60} & \cellcolor{top3}{0.3300}   \\
& \new{CoR-GS}~\cite{zhang2025cor}         & \cellcolor{top3}{0.3850}   & \cellcolor{top3}{12.82}  & \cellcolor{top3}{0.5550}       & 0.4925   & \cellcolor{top3}{14.90}  & \cellcolor{top3}{0.4075}        & \cellcolor{top3}{0.5950} & 18.00 & 0.3725   \\
& \new{Instantsplat}~\cite{fan2024instantsplat} & \cellcolor{top2}{0.4025}   & \cellcolor{top2}{13.65}  & \cellcolor{top2}{0.5425}       & \cellcolor{top3}{0.5300}   & \cellcolor{top2}{16.46}  & \cellcolor{top2}{0.3700}    & \cellcolor{top2}{0.6050}& \cellcolor{top2}{19.28} & \cellcolor{top2}{0.3450}   \\
% \cmidrule{2-11}
& LM-Gaussian    & \cellcolor{top1}{0.4600}   & \cellcolor{top1}{14.68}  & \cellcolor{top1}{0.4725}         & \cellcolor{top1}{0.6205}  & \cellcolor{top1}{18.40} & \cellcolor{top1}{0.2412}      & \cellcolor{top1}{0.6875} & \cellcolor{top1}{20.54}  & \cellcolor{top1}{0.2300}   \\
\midrule
\multirow{6}{*}[-1em]{\rotatebox[origin=c]{90}{MipNeRF360}} 
& FreeNeRF~\cite{yang2023freenerf}   & 0.2575   & 9.92  & 0.7250           & 0.2950 & 11.67  & 0.6275        & 0.3275    & 14.86   & 0.5500   \\
& SparseNerf~\cite{wang2023sparsenerf}    & 0.2850 & 10.06  & 0.7075          & 0.3050 & 11.78  & 0.6400        & 0.3525    & 15.11   & 0.5150   \\ 
& DNGaussian~\cite{li2024dngaussian} & 0.3375   & 11.14  & 0.6375          & 0.3525 & 12.46  & 0.6550        & 0.3775    & 15.96   & 0.4800   \\
& \new{Scaffold-GS}~\cite{lu2024scaffold}  &0.3250   & 11.92  & 0.6550          & 0.3225 & 14.30  & \cellcolor{top3}{0.5525}        & 0.4325    & 18.25   & 0.3825 \\
& \new{Splatfield}~\cite{mihajlovic2025splatfields} & 0.3475 & 10.52& \cellcolor{top3}{0.6175}     & 0.3250 & 13.41  & 0.5700        & 0.4425    & 17.49   & 0.4225   \\
& \new{CoR-GS}~\cite{zhang2025cor}         & \cellcolor{top2}{0.4025}   & \cellcolor{top2}{14.55}  & 0.6675          & \cellcolor{top3}{0.3925} & \cellcolor{top3}{15.75}  & 0.5975        & \cellcolor{top3}{0.4975}    & \cellcolor{top2}{18.60}   & \cellcolor{top3}{0.3575}   \\
& \new{Instantsplat}~\cite{fan2024instantsplat} & \cellcolor{top3}{0.4025}   & \cellcolor{top3}{14.41}  & \cellcolor{top2}{0.5450}     & \cellcolor{top2}{0.4700} & \cellcolor{top2}{16.57}  & \cellcolor{top2}{0.4125}        & \cellcolor{top2}{0.5125}    & \cellcolor{top3}{18.33}   & \cellcolor{top2}{0.3425}   \\
% \cmidrule{2-11}
& LM-Gaussian    & \cellcolor{top1}{0.4400} & \cellcolor{top1}{15.18}  & \cellcolor{top1}{0.5350}    & \cellcolor{top1}{0.5475}  & \cellcolor{top1}{17.49} & \cellcolor{top1}{0.3300}        & \cellcolor{top1}{0.5800}   & \cellcolor{top1}{19.22}   & \cellcolor{top1}{0.3000}   \\
\bottomrule
\end{tabular}}
\end{table*}

\begin{figure*}
\includegraphics[width=\textwidth]{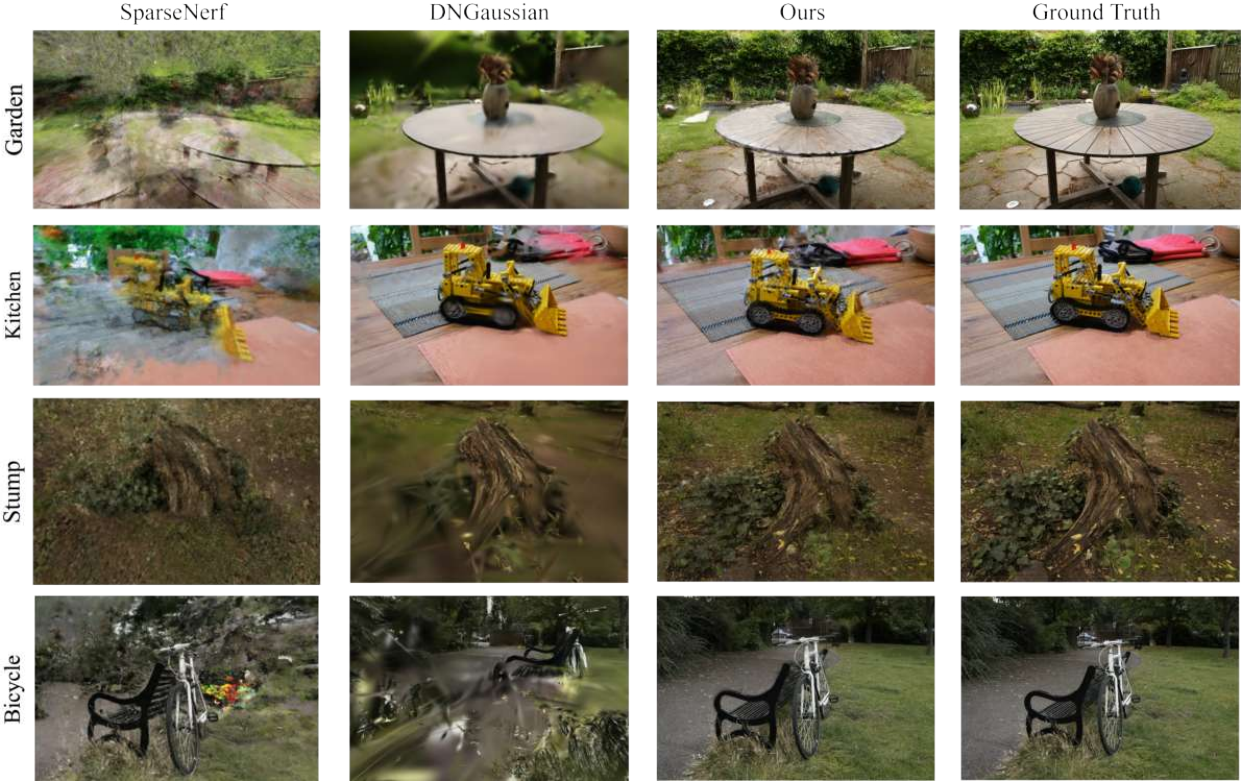}
\caption{\textbf{Qualitative comparison on MipNerf360 Dataset with 16 input views.} Similar to Tanks and Temple Dataset, our approach consistently fairs better in recovering image structure from foggy geometry, where baselines typically struggle
with floaters and artifacts.} \label{mipresult}
\end{figure*}
\subsection{Experimental Setup}
\boldparagraph{Dataset} Our experiments were conducted using three datasets: the Tanks and Temples Dataset~\cite{knapitsch2017tanks}, the MipNeRF360 Dataset~\cite{barron2022mip}, and the LLFF Dataset~\cite{mildenhall2019local}. The Tanks and Temples and MipNeRF360 datasets feature 360-degree real-world scenes, while the LLFF dataset comprises feed-forward scenes. From the Tanks and Temples Dataset, we uniformly selected 200 images covering scenes like Family, Horse, Ignatius, and Trunk to represent the entire 360-degree environments. In the MipNeRF360 dataset, we chose the initial 48 frames capturing various elevations across a full 360-degree rotation, including scenes such as Garden, Bicycle, Kitchen, and Stump. Additionally, scenes like flowers, orchids, and ferns from the LLFF Dataset have been incorporated as well.

% \boldparagraph{Train/Test Datasets Split} For 360-degree scenes, we varied the view number $K$ from 4 to 16 to assess all the algorithms under consideration. In a training set comprising $K$ images, the remaining images were allocated to the test set. Concerning the feed-forward dataset, we adhered to the setup outlined in previous research \cite{li2024dngaussian, niemeyer2022regnerf, wang2023sparsenerf}, employing 3 views for training and the remainder for testing.

\boldparagraph{Metrics}  In the assessment of novel view synthesis, we present Peak Signal-to-Noise Ratio (PSNR), Structural Similarity Index Measure (SSIM)\cite{wang2004image}, and Learned Perceptual Image Patch Similarity (LPIPS)\cite{zhang2018unreasonable} scores as quantitative measures to evaluate the reconstruction performance.

\boldparagraph{Baselines} We compare our method against \old{\st{7}}\new{11} baseline approaches. Our evaluation included sparse-view reconstruction methods such as DNGaussian~\cite{li2024dngaussian}, FreeNeRF~\cite{yang2023freenerf}, SparseNeRF~\cite{wang2023sparsenerf}, PixelNeRF~\cite{yu2021pixelnerf}, MVSNeRF~\cite{chen2021mvsnerf}, DietNeRF~\cite{jain2021putting}, RegNerf~\cite{niemeyer2022regnerf}, Scaffold-GS~\cite{lu2024scaffold}, Splatfield~\cite{mihajlovic2025splatfields}, CoR-GS~\cite{zhang2025cor} and InstantSplat~\cite{fan2024instantsplat}. Additionally, we compared our method with the vanilla 3DGS approach to evaluate our scalability.

\subsection{Implementation Details}
\label{Implementation}
We implemented our entire framework in PyTorch 2.0.1 and conducted all experiments on an A6000 GPU. In the Background-Aware Depth-guided Initialization stage, 
% we performed joint optimization of aligned point clouds over 300 iterations. 
loss weights $\alpha_g$, $\alpha_s$ and $\alpha_l$ were set to 0.01, 0.01 and 0.1. Geometry-based cleaning was applied every 50 iterations. For Confidence-based cleaning, $\tau_1$ and $\tau_2$ were set to 0.1 and 0, respectively. Moving on to the Multi-modal Regularized Gaussian Reconstruction stage, we trained the Gaussian model for 6,000 iterations with specified loss function weights: $\beta_{vir}$ = 0.5, $\beta_{dep}$ = 0.3, and $\beta_{nor}$ = 0.1 across all experiments. We use ControlNet~\cite{zhang2023adding} as our foundational model. 
% During the training of the Gaussian Repair Model, all images in the training datasets were resized to 512$\times$512 to match ControlNet's input dimensions. 
The low-rank $r$ was set to 64, and the model was trained for 2000 steps. In the Iterative Gaussian Refinement Module, the denoising strength was set at 0.3 for each repair iteration, and the repair process was repeated every 4,000 iterations. The value of $\beta_{rep}$ was adjusted based on the distance between the virtual view and its nearest training view, within the range of (0, 1). This iterative cycle was set to 3 in our experiments. 

\subsection{Quantative and Qualitative Results} 

\begin{figure*}
\includegraphics[width=\textwidth]{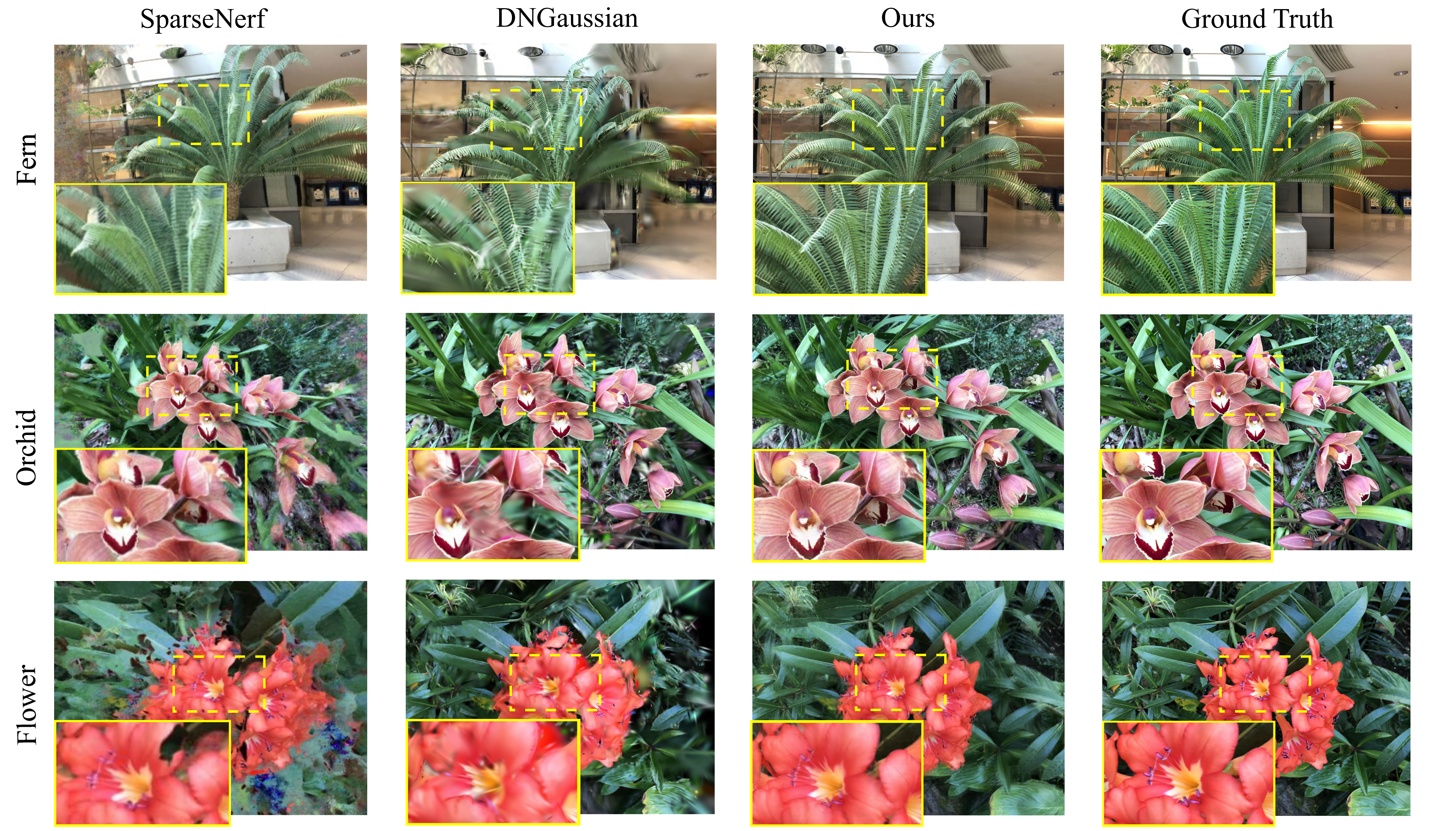}
\caption{\textbf{Qualitative comparison on LLFF Dataset with 3 input views.} Compared to baseline methods like DNGaussian, FreeNerf, and SparseNerf, our technique delivers enhanced results and greater detail, as demonstrated by PSNR, SSIM, and LPIPS scores.} \label{llff}
\end{figure*}

\begin{figure*}
\includegraphics[width=\textwidth]{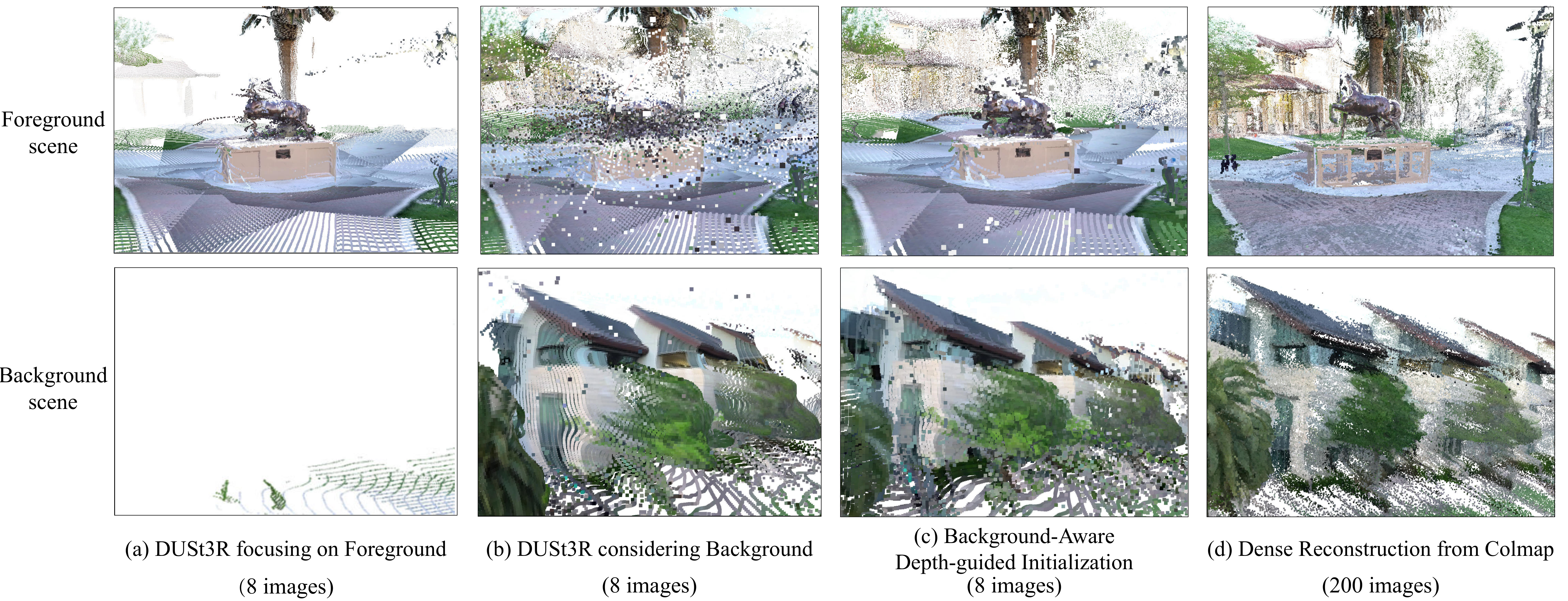}
\centering
\caption{\textbf{Point Cloud visualizations of DUSt3R and Background-Aware Depth-Guided Initialization.} As shown in images (a) and (b), with different confidence threshold for cleaning, DUSt3R either suffer from the empty or significant artifacts in the background part. Through the utilization of Depth-Guided Optimization, Point Cloud Cleaning, and Foreground-Background Separation techniques, our module excels in producing enhanced point clouds while addressing issues such as floaters and scene distortion. Image (d) displays the dense reconstruction result from 200 images by Colmap, serving as a useful reference.} \label{pointcloud}
\end{figure*}

% \boldparagraph{Test poses estimation by incremental adding and refinement} Because ground truth poses are unknown in our framework and poses recovered from Background-Aware Depth-guided Initialization don't align with those got from Colmap. To test the quality of reconstruction, such as calculating PSNR and LPIPS, we need to get test poses set in advance. Here we designed a direct and efficient approach to solve the problem. Assuming we have $K$ train views and $L$ test views (e.g., $K$ = 8, $L$ = 100), we first use our Background-Aware Depth-guided Initialization module to recover $K$ train-view poses from input images. Considering memory consumption, we adopt an incremental strategy by adding $Q$ (e.g., $Q$ = 16) test views into the system once a time. We randomly pick $Q$ test views and form a batch ($K$ train views and $Q$ test views). This batch will be put into the Initialization Module again as former. The difference is that those $K$ train poses are fixed as ground-truth and only test poses will be optimized to align to train-view coordinates.\par
% Obviously, test poses got by this incremental adding are not precise, which may cause bias into the evaluation system. Here we use a pose refinement module to achieve a more precise alignment and rendering of the test views for equitable comparisons. Specifically, we maintain the trained scene model in a frozen state while only optimizing parameters of the camera poses for test views. This optimization process focuses on minimizing the photo-metric discrepancies between the synthesized images and the actual test views.

\boldparagraph{Tanks and Temples \& MipNerf360} \new{The quantitative results presented in Table \ref{tab: finalresult} demonstrate that our method consistently outperforms others in terms of PSNR, SSIM, and LPIPS metrics across various input views. Visual results are also showcased in Figure \ref{tntresult} and Figure \ref{mipresult}, where our method preserves more structures and finer details. We attribute LM-Gaussian's outstanding performance to three main factors. First, instead of relying on colmap initialization like other methods~\cite{yang2023freenerf,wang2023sparsenerf,li2024dngaussian,mihajlovic2025splatfields,lu2024scaffold,zhang2025cor}, we incorporate stere prior into our model, maintaining robustness in sparse-view settings where traditional SfM methods struggle to provide reliable point clouds and camera poses. Second, we employ customizable regularization strategies to prevent over-fitting, similar to frequency~\cite{yang2023freenerf}, depth~\cite{wang2023sparsenerf,li2024dngaussian} and correlation~\cite{zhang2025cor} regularizations adopted in other sparse-view reconstruction methods. Third, we introduce generative model priors to help restore scene details,  a feature lacking in other methods~\cite{fan2024instantsplat, lu2024scaffold, mihajlovic2025splatfields}.}

% \begin{table}[!t]
% \caption{\textbf{Quantitative Comparison on LLFF Dataset for 3 input views.} The best, second-best, and third-best entries are marked in red, yellow, and green, respectively. Our method shows the best reconstruction result compared with other sparse-view reconstruction works.}
% \label{tab:llffdtu}
% \setlength{\abovecaptionskip}{4pt}
% \resizebox{1\linewidth}{!}{
% \setlength{\tabcolsep}{3.2 mm}
% \begin{tabular}{l|ccc}
% \hline
% \multirow{2}{*}{Methods} & \multicolumn{3}{c}{LLFF}          \\    
%  & PSNR $\uparrow$    & LPIPS $\downarrow$   & SSIM $\uparrow$     \\ \hline
% PixelNeRF~\cite{yu2021pixelnerf} & 15.17   & 0.612   & 0.338   \\
% MVSNeRF~\cite{chen2021mvsnerf}  & 16.88  & 0.427 & 0.484 \\ 
% DietNeRF~\cite{jain2021putting}  & 14.94    & 0.496   & 0.370   \\
% RegNeRF~\cite{niemeyer2022regnerf} & 18.08   & 0.396    & 0.487   \\
% FreeNeRF~\cite{yang2023freenerf}  & 18.63 & 0.328 & 0.512 \\      
% \new{SparseNerf}~\cite{wang2023sparsenerf}  &\new{18.52}  & \new{0.335} & \new{0.527}   \\
% DNGaussian~\cite{li2024dngaussian}  & 18.32    & 0.314   & 0.535  \\
% Scaffold-GS~\cite{lu2024scaffold}  & 18.32    & 0.314   & 0.535  \\
% Splatfield  & 19.28    & 0.266   & 0.607 \\
% CoR-GS~\cite{zhang2025cor}  & 18.91    & 0.309   & 0.594  \\
% Instantsplat  & 19.33    & 0.242   &0.628   \\\hline
% LM-Gaussian  & \cellcolor{top1}{19.63} & \cellcolor{top1}{0.228} & \cellcolor{top1}{0.644} \\  \hline
% \end{tabular}                                                       
% }
% \vspace{-.4cm}
% \end{table}

\begin{table}[!t]
\caption{\textbf{Quantitative Comparison on LLFF Dataset for 3 input views.} \new{Best results are highlighted as \colorbox{top1}{1st}, \colorbox{top2}{2nd} and \colorbox{top3}{3rd}}. Our method shows the best reconstruction result compared with other sparse-view reconstruction works.}
\label{tab:llffdtu}
\setlength{\abovecaptionskip}{4pt}
\resizebox{1\linewidth}{!}{
\setlength{\tabcolsep}{3.2 mm}
\begin{tabular}{l|ccc}
\hline
\multirow{2}{*}{Methods} & \multicolumn{3}{c}{LLFF}          \\    
 & PSNR $\uparrow$    & LPIPS $\downarrow$   & SSIM $\uparrow$     \\ \hline
PixelNeRF~\cite{yu2021pixelnerf} & 15.17   & 0.612   & 0.338   \\
MVSNeRF~\cite{chen2021mvsnerf}  & 16.88  & 0.427 & 0.484 \\ 
DietNeRF~\cite{jain2021putting}  & 14.94    & 0.496   & 0.370   \\
RegNeRF~\cite{niemeyer2022regnerf} & 18.08   & 0.396    & 0.487   \\
FreeNeRF~\cite{yang2023freenerf}  & 18.63 & 0.328 & 0.512 \\      
\new{SparseNerf}~\cite{wang2023sparsenerf}  &18.52 & 0.335 & 0.527   \\
DNGaussian~\cite{li2024dngaussian}  & 18.32    & 0.314   & 0.535  \\
\new{Splatfield}~\cite{mihajlovic2025splatfields}   & 17.94    & 0.402   & 0.499  \\
\new{Scaffold-GS}~\cite{lu2024scaffold} & 18.88    & 0.309   & 0.567 \\
\new{CoR-GS}~\cite{zhang2025cor}  & \cellcolor{top3}{18.91}    & \cellcolor{top3}{0.292}   & \cellcolor{top3}{0.594}  \\
\new{Instantsplat}~\cite{fan2024instantsplat}  & \cellcolor{top2}{19.33}    & \cellcolor{top2}{0.242}   &\cellcolor{top2}{0.628}   \\\hline
LM-Gaussian  & \cellcolor{top1}{19.63} & \cellcolor{top1}{0.228} & \cellcolor{top1}{0.644} \\  \hline
\end{tabular}                                                       
}
\vspace{-.4cm}
\end{table}

\boldparagraph{LLFF} Besides challenging 360-degree large-scale scenes, we also conducted experiments on feed-forward scenes like the LLFF Dataset to ensure the thoroughness of our study and validate the robustness of our method. Following previous sparse-view reconstruction works, we take 3 images as input, with quantitative results presented in Table \ref{tab:llffdtu} and qualitative results shown in Figure \ref{llff}. It is observed that sparse-view methods like DNGaussian also demonstrate commendable visual outcomes and relatively high PSNR and SSIM values. This can be attributed to the nature of the LLFF Dataset, which does not encompass a 360-degree scene but rather involves movement within a confined area. This results in higher image overlap and fewer unobserved areas, making it easier to reconstruct the scene. However, despite the impressive performance of these methods, our approach still exhibits certain advantages. In addition to the numerical enhancements demonstrated in TABLE \ref{tab:llffdtu}, taking the flower scene as an example, our method excels in visual results by restoring finer details such as flower textures. Moreover, our method maintains superior performance in regions with less overlap, as exemplified by the leaves in the surroundings.

\begin{figure*}
\includegraphics[width=\textwidth]{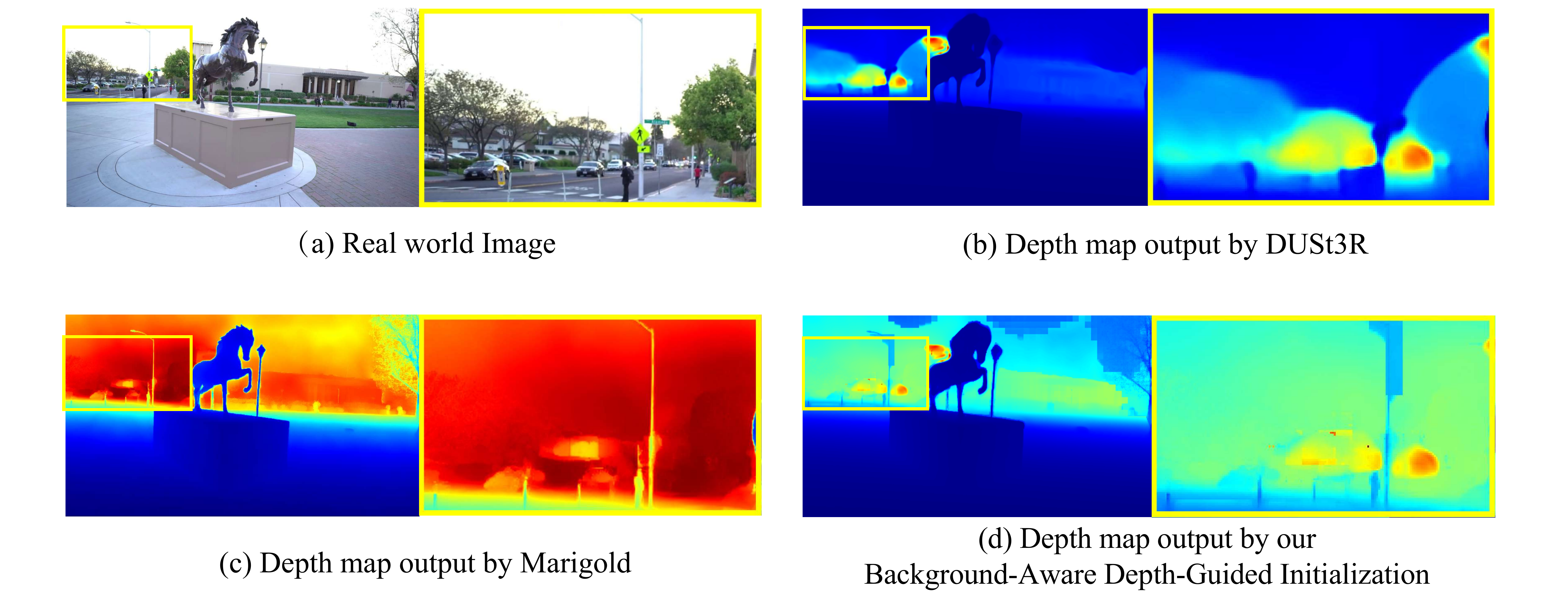}
\caption{\textbf{Comparison of Point Cloud Depth Before and After Depth-Guided Optimization.} The depth maps generated by DUSt3R display a blurred background for the input image. In contrast, by leveraging depth priors, our Background-Aware Depth-Guided Initialization produces significantly enhanced depth maps, offering a more precise representation of the scene's depth. While the optimized depth may still exhibit some imperfections, particularly around the streetlights, these issues will be addressed in further refinement stages.}\label{depth}
\end{figure*}

\subsection{Ablation Study} 
\label{ab}
\begin{figure}
\includegraphics[width=0.5\textwidth]{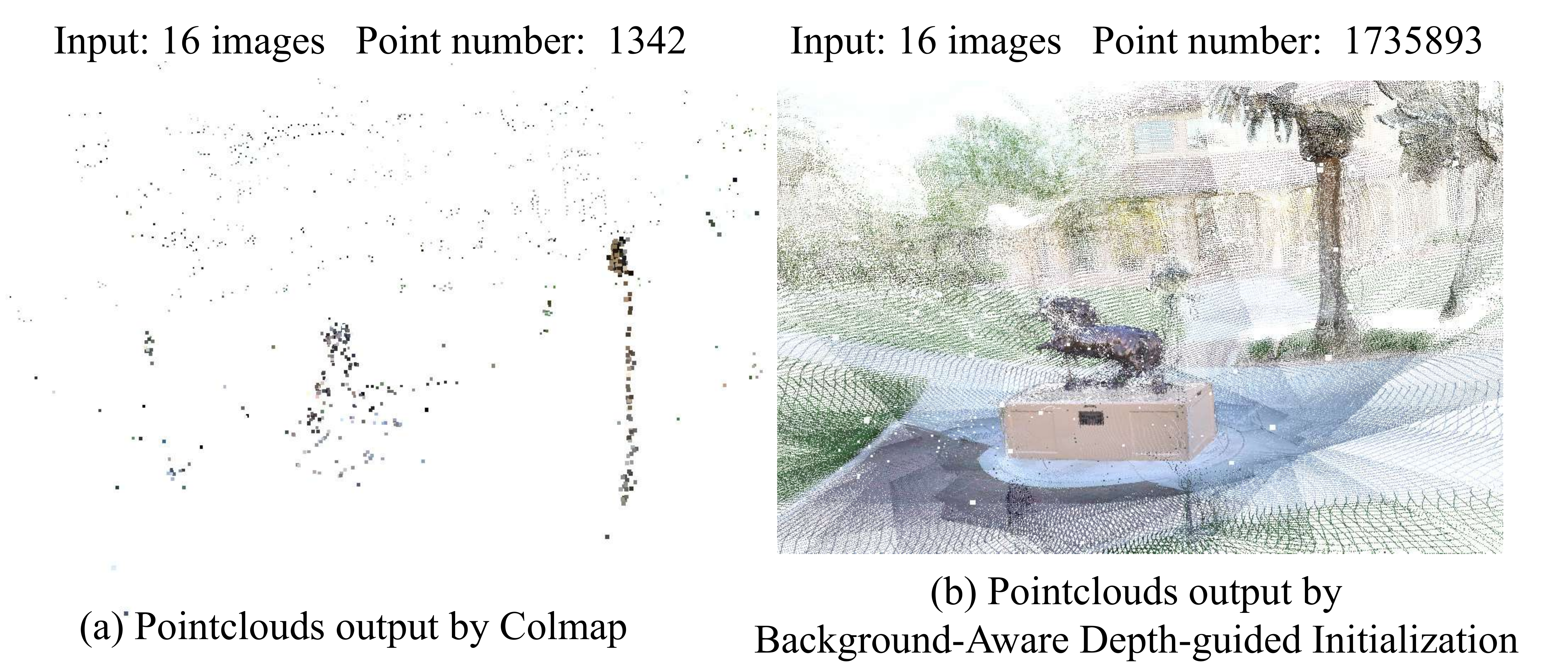}
\caption{Point clouds output by Colmap and Background-Aware Depth-guided Initialization. (a) Colmap fail to reconstruct reliable 3d points. (b) Dense pointclouds are obatined by Multi-modal Prior-guided Initialization } \label{Colmap}
\end{figure}

\boldparagraph{Colmap Initialization} Initially, we investigated the conventional Colmap method within our sparse-view settings. It fails to reconstruct point clouds with 8 input images in 360-degree scenes. We progressively increased the number of input images until Colmap could eventually generate sparse point clouds. With 16 input images, as illustrated in Figure \ref{Colmap}, Colmap's resulting point clouds were significantly sparse, comprising only 1342 points throughout the scene. In contrast, our Background-Aware Depth-guided Initialization method excels in generating high-quality dense point clouds.

\begin{figure*}
\includegraphics[width=\textwidth]{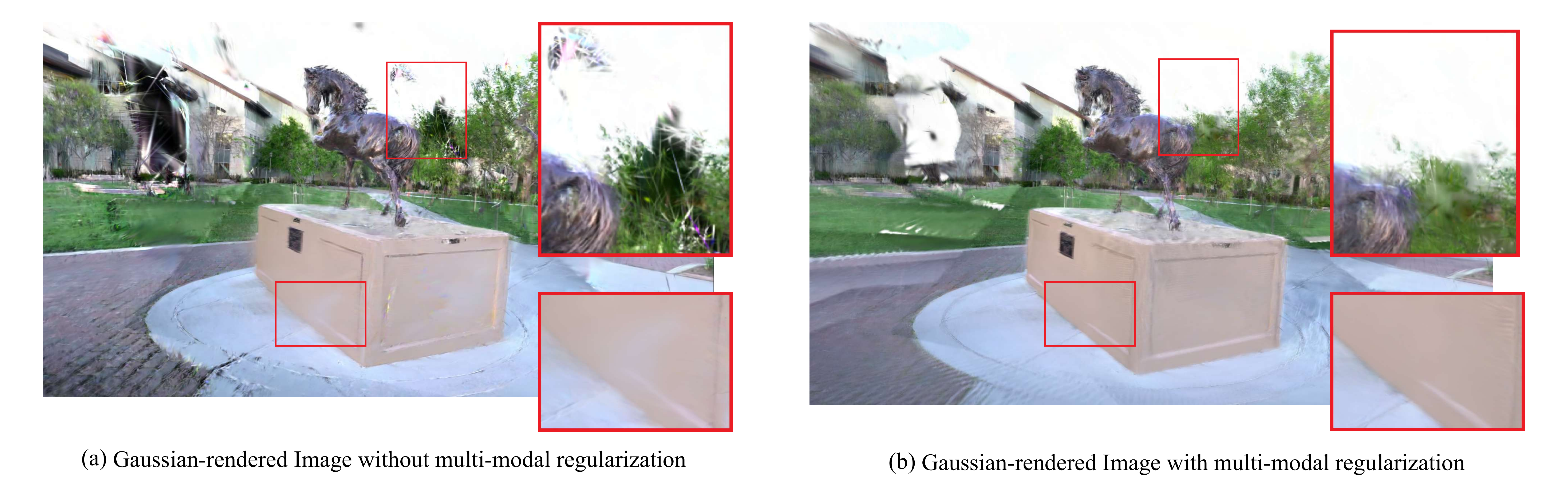}
\caption{\textbf{Qualitative Comparison between Gaussian-rendered images with and without multi-modal regularization.} Through multi-modal regularization, Gaussian-rendered images exhibit smoother surfaces and reduced artifacts within the scene. In contrast, images lacking regularization display black holes on houses, trees, and sharp angles on the ground that detract from the overall quality.}  \label{regular}
\end{figure*}

\begin{table}[!t]
\caption{\textbf{Ablation Study of Different Initializations.} We compare our Background-Aware Depth-guided Initialization with Colmap and DUSt3R on the Horse scene with 16-view setting. }\label{tab:initial}
\resizebox{1\linewidth}{!}{
\centering
\renewcommand{\arraystretch}{1.5} % 调整行高
\begin{tabular}{c|c|c|c}
\hline
Initialization  & PSNR $\uparrow$ & LPIPS $\downarrow$ &SSIM $\uparrow$     \\ \hline
Colmap  & 13.42 & 0.558 & 0.192   \\ \hline
DUSt3R  & 17.12 & 0.328 & 0.546   \\ \hline
Proposed Initialization & \textbf{18.04}  & \textbf{0.304}   & \textbf{0.576}     \\  \hline
\end{tabular}}
\end{table}

\boldparagraph{Effect of Background-Aware Depth-guided Initialization}
\label{sec:withDUSt3R}
% Our proposed Background-Aware Depth-guided Initialization excels in managing sparse-view input settings and reconstructing dense point clouds and camera poses. 
Through visual demonstrations, we highlight DUSt3R's limitations in reconstructing high-quality background scenes, plagued by artifacts and distortions. As depicted in Figure \ref{pointcloud}(a)(b), DUSt3R either lacks background details or presents poor background quality, resulting in subpar reconstructions.
% In Figure \ref{pointcloud}(a), using a high confidence threshold for point cloud cleaning encourages DUSt3R to prioritize foreground object reconstruction, neglecting a holistic background representation. Conversely, using a lower confidence threshold, as depicted in Figure \ref{pointcloud}(b), DUSt3R's point cloud being riddled with artifacts and floaters, culminating in inaccurate and noisy Gaussian initialization. Furthermore, evident distortion issues in background scenes stem from the absence of multi-view supervision and scale error correction.

In contrast, as illustrated in Figure \ref{pointcloud}~(c), our module adeptly reconstructs background scenes with minimal distortion while preserving high-quality foreground scenes, exhibiting fewer artifacts and floaters compared to Figure (b)'s point clouds. Additionally, we present visualizations of depth maps before and after depth optimization. In Figure \ref{depth}, the original DUSt3R's depth maps reveal background blurriness, blending elements like the sky and street lamps. With guidance from Marigold, our Background-Aware Depth-guided Initialization notably enhances the reconstruction of background scenes compared to the original DUSt3R output.

Moreover, Table \ref{tab:initial} provides quantitative comparisons among our method, Colmap, and DUSt3R. While Colmap yields less favorable results, DUSt3R shows significant improvement. Despite DUSt3R's performance, our initialization module achieves state-of-the-art outcomes, leading to improvements in PSNR and SSIM metrics.

\boldparagraph{Effect of Multi-modal regularized Gaussian Reconstruction}
We conducted ablation studies on the multi-modal regularization, incorporating depth, normal, and virtual-view regularization. The quantitative outcomes are detailed in Table \ref{tab:ablate}. Following the integration of these regularization techniques, the novel view synthesis demonstrates improved results, indicated by higher PSNR and SSIM values, as well as finer details with lower LPIPS scores. With the implementation of multi-modal regularization, as depicted in Figure \ref{regular}, the Gaussian-rendered images showcase smoother surfaces and reduced artifacts within the scene. Conversely, images lacking regularization exhibit black holes and sharp angles, diminishing the overall quality.
\begin{figure*}
\includegraphics[width=\textwidth]{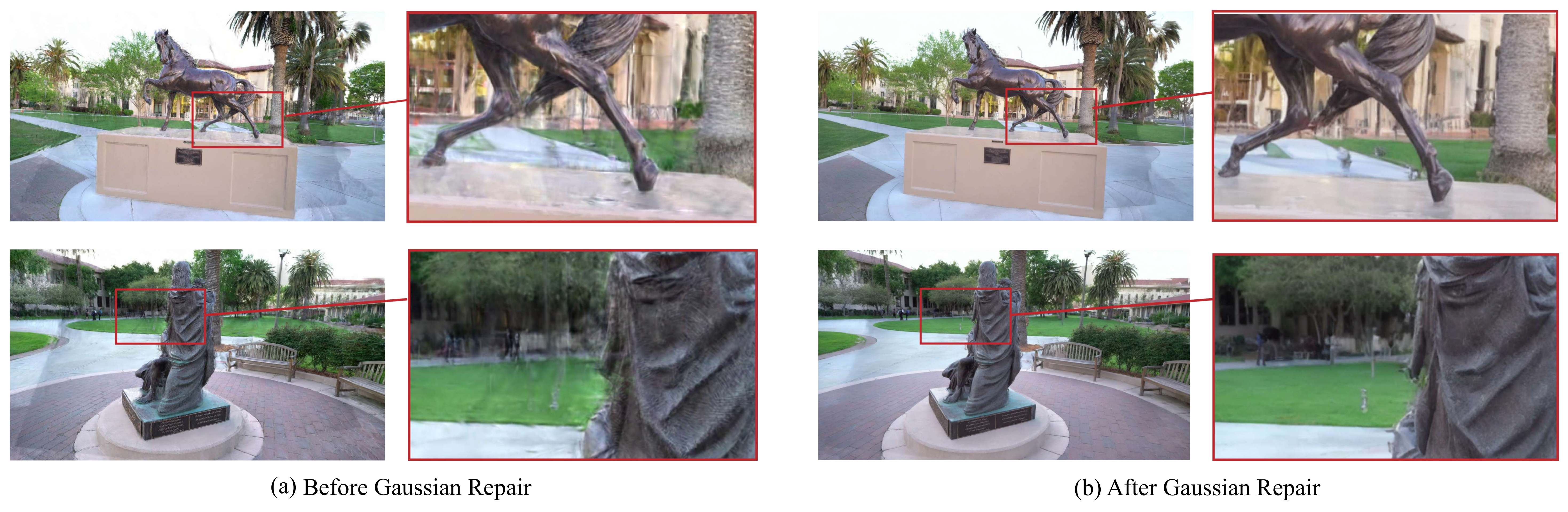}
\caption{\textbf{Visual Comparison of Images before and after Gaussian Repair}: The images displayed on the left showcase Gaussian-rendered images derived from Coarse Gaussian Reconstruction Module. Noticeably, these images exhibit blurriness and artifacts. In contrast, the images on the right demonstrate a marked improvement after being repaired by our Gaussian Repair Model, showcasing a cleaner and higher-quality outcome. } \label{repairvis}
\end{figure*}

\begin{table}[!t]
\caption{\textbf{Ablation Study.} We ablate our method on the Horse scene with 16-view setting.  Background-Aware Depth-guided Initialization(BA), Regularization strategies and Iterative Gaussian Refinement all improves the novel view synthesis quality.}\label{tab:ablate1}
\resizebox{1\linewidth}{!}{
\centering
\renewcommand{\arraystretch}{1.5} % 调整行高
\begin{tabular}{ccc|ccc}
\hline
\multicolumn{3}{c}{Method} & \multicolumn{3}{c}{Metric}  \\
BA  & Regularization & Refinement & PSNR $\uparrow$ & LPIPS $\downarrow$ &SSIM $\uparrow$     \\ \hline
$\times$ & $\times$ & $\times$ & 13.42 & 0.558 & 0.192   \\ \hline  
$\checkmark$ & $\times$ & $\times$ & 18.04 & 0.304  & 0.576   \\ \hline
$\checkmark$ & $\checkmark$ & $\times$ & 21.32 & 0.145 & 0.731        \\ \hline
$\checkmark$ & $\checkmark$ & $\checkmark$ & \textbf{22.04}  & \textbf{0.119}   & \textbf{0.776}     \\  \hline
\end{tabular}}
\end{table}

\begin{table}[!t]
\caption{\textbf{Regularization Test.} We individually test the multi-depth regularization, cosine-normal regularization and weighted point-render regularization }\label{tab:ablate}
\resizebox{1\linewidth}{!}{
\centering
\renewcommand{\arraystretch}{1.5} % 调整行高
\begin{tabular}{ccc|ccc}
\hline
\multicolumn{3}{c}{Regularizations} & \multicolumn{3}{c}{Metric}  \\
Depth  & Normal & Virtual-view & PSNR $\uparrow$ & LPIPS $\downarrow$ &SSIM $\uparrow$     \\ \hline
$\times$ & $\times$ & $\times$ & 18.04 & 0.304  & 0.576   \\ \hline  
$\checkmark$ & $\times$ & $\times$ & 19.74 & 0.205  & 0.634   \\ \hline
$\checkmark$ & $\checkmark$ & $\times$ & 20.02 & 0.188 & 0.665        \\ \hline
$\checkmark$ & $\checkmark$ & $\checkmark$ & \textbf{21.32}  & \textbf{0.145}   & \textbf{ 0.731}     \\  \hline
\end{tabular}}
\end{table}

\boldparagraph{Effect of Iterative Gaussian Refinement} We further explore the usefulness of the iterative Gaussian refinement module. As illustrated in Figure \ref{repairvis}, we present a comparison between the images before and after Gaussian Repair. The noticeable outcomes highlight that the repaired images exhibit enhanced details and a reduction in artifacts, emphasizing the effectiveness of the Gaussian Repair Model. Quantitative results before and after the Iterative Gaussian Refinement are also detailed in Table \ref{tab:ablate1}. While a marginal improvement in PSNR is noted, more intricate metrics such as LPIPS and SSIM demonstrate substantial enhancements. These findings align seamlessly with our primary objective of restoring intricate details within the images.

% \textbf{Why not directly apply the diffusion model} Using the stable diffusion model directly for scene restoration is not a good choice. This is due to the model's tendency to produce diverse outputs and undergo a highly random restoration process. The generated images frequently lack the stylistic and content consistency required for the restoration task. Not only does this inconsistency fail to contribute to the reconstruction process, but it can also perplex the Gaussian network, causing convergence problems and potentially resulting in an overall blurry outcome. \lina{this part should be placed in EXP, and add visual results or quantitative results. }

\begin{figure}
\includegraphics[width=0.5\textwidth]{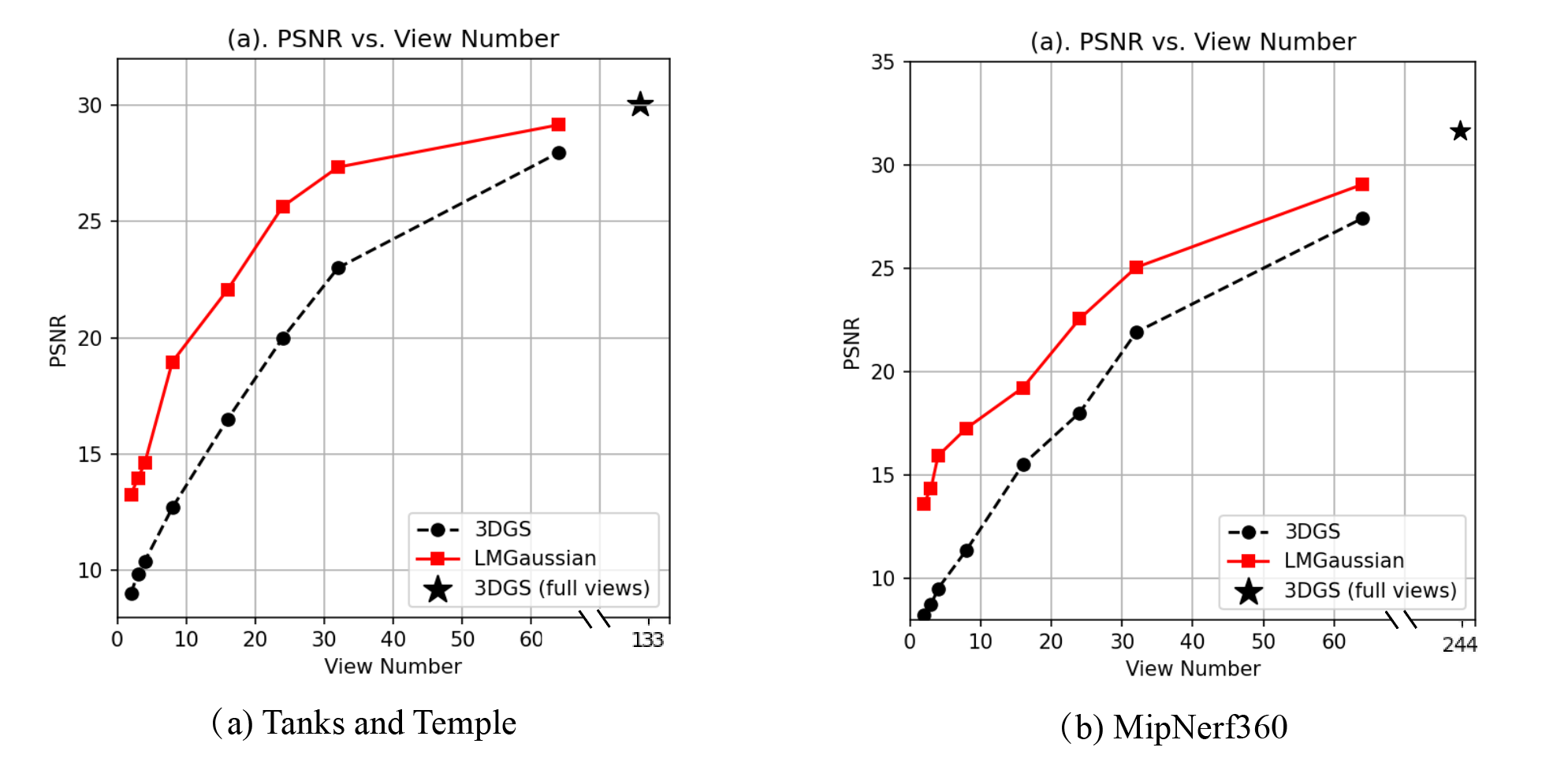}
\caption{\new{Scalability of LM-Gaussian with input views. LM-Gaussian demonstrates superior performance compared to the vanilla 3DGS, even over 32 input views.}} \label{views}
\end{figure}

\boldparagraph{The number of input images} In Figure \ref{views}, we assess our method using different number of sparse input images $N$. We compare LM-Gaussian with the original 3DGS across view splits of growing sizes $K \in \{\new{2, 3}, 4, 8, 16, 24, 32\new{, 64\}}$ in the Tanks and Temples \new{and MipNerf360} Dataset. Notably, \new{even in over 32 view images, our method still shows a better performance than 3DGS}.

\section{Conclusions}
We introduce LM-Gaussian, a sparse-view 3D reconstruction method that harnesses priors from large vision models. Our method includes a robust initialization module that utilizes stereo priors to aid in recovering camera poses and reliable Gaussian spheres. Multi-modal regularizations leverage monocular estimation priors to prevent network overfitting. Additionally, we employ iterative diffusion refinement to incorporate extra image diffusion priors into Gaussian optimization, enhancing scene details. Furthermore, we utilize video diffusion priors to further improve the rendered images for realistic visual effects. Our approach significantly reduces the data acquisition requirements typically associated with traditional 3DGS methods and can achieve high-quality results even in 360-degree scenes. LM-Gaussian currently is built on standard 3DGS that only works well on static scenes, and we would like incorporate dynamic 3DGS techniques to enable dynamic modeling in the future.

{

\bibliographystyle{plain}
\bibliography{main}

}

\vfill

\end{document}